%% file: main.tex
\def\year{2021}\relax
\documentclass[letterpaper]{article} 
\usepackage{aaai21}  
\usepackage{times}  
\usepackage{helvet} 
\usepackage{courier}  
\usepackage[hyphens]{url}  
\usepackage{graphicx} 
\usepackage{natbib}  
\usepackage{caption} 
\usepackage{comment}
\usepackage{amsmath,amssymb} 
\usepackage{color}
\usepackage{colortbl}
\usepackage{float}
\usepackage[linesnumbered,ruled,vlined,noend]{algorithm2e}
\usepackage{multirow}
\usepackage{soul}
\usepackage{enumitem}
\usepackage{hyperref}

\newtheorem{definition}{Definition}

\newcommand{\Lpers}{\mathcal{L}_{Pers}}
\newcommand{\Ldice}{\mathcal{L}_{DICE}}
\newcommand{\calM}{\mathcal{M}}
\DeclareMathOperator{\pers}{Pers}

\definecolor{Gray}{gray}{0.92}
\definecolor{LightCyan}{rgb}{0.88,1,1}

\DeclareMathOperator{\visited}{visited}
\DeclareMathOperator{\birth}{birth}

\DeclareMathOperator{\argmax}{argmax}

\newif\ifdraft
\drafttrue

\newcommand{\myparagraph}[1]{\smallskip\noindent\textbf{#1}}

\title{Localization in the Crowd with Topological Constraints}
\author{
    Shahira Abousamra\textsuperscript{\rm 1},\;  
    Minh Hoai\textsuperscript{\rm 1},\;
    Dimitris Samaras\textsuperscript{\rm 1},\;
    Chao Chen\textsuperscript{\rm 2}
    \\
}
\affiliations{

    \textsuperscript{\rm 1}Department of Computer Science, Stony Brook University,  USA \\
    \textsuperscript{\rm 2}Department of Biomedical Informatics, Stony Brook University, USA \\
    \{sabousamra, minhhoai, samaras\}@cs.stonybrook.edu, chao.chen.1@stonybrook.edu

}

\begin{document}
\input{definitions}

\maketitle

\begin{abstract}
We address the problem of crowd localization, i.e., the prediction of dots corresponding to people in a crowded scene. 
Due to various challenges, a localization method is prone to spatial semantic errors, i.e., predicting multiple dots within a same person or collapsing multiple dots in a cluttered region.
We propose a topological approach targeting these semantic errors. We introduce a topological constraint that teaches the model to reason about the spatial arrangement of dots.
To enforce this constraint, we define a persistence loss based on the theory of persistent homology. 
The loss compares the topographic landscape of the likelihood map and the topology of the ground truth.
Topological reasoning improves the quality of the localization algorithm especially near cluttered regions. 
On multiple public benchmarks, our method outperforms previous localization methods. 
Additionally, we demonstrate the potential of our method in improving the performance in the crowd counting task.\footnote{The code and a full version of this paper can be found at \newline \url{https://github.com/TopoXLab/TopoCount}.}

\end{abstract}

\section{Introduction}
\label{sec:intro}
\input{sec-intro}

\section{Related Work}
\label{sec:relatedwork}
\input{sec-relatedwork}

\section{Method: TopoCount}
\label{sec:method}
\input{sec-method}


\input{sec-experiments-pre}
\section{Experiments}
\label{sec:experiments}
\input{sec-experiments-post}

\section{Conclusion}
This paper proposes a novel method for localization in the crowd. We propose a topological constraint and a novel persistence loss based on persistent homology theory.  
The proposed topological constraint is flexible and suitable for both sparse and dense regions. 
The proposed method achieves state-of-the-art localization accuracy.
The high quality of our results is further demonstrated by the significant boost of the performance of density-based counting algorithms when using our results as additional input.
Our method closes the gap between the performance of localization and density map estimation methods; thus paving the way for advanced spatial analysis of crowded scenes in the future.

\subsection*{Acknowledgements}
This research was partially supported by US National Science Foundation Awards IIS-1763981, CCF-1855760, IIS-1909038,
the SUNY2020 Infrastructure Transportation Security Center, Air Force Research Laboratory
(AFRL) DARPA FA8750-19-2-1003, the Partner University Fund, and a gift from Adobe.
We thank anonymous
reviewers for constructive comments and suggestions.
 \clearpage

\bibliography{egbib,topo-bib}
\clearpage

\input{sec-supplemental}

\clearpage

\end{document}

%% file: definitions.tex
\def\mA{\mathcal{A}}
\def\mB{\mathcal{B}}
\def\mC{\mathcal{C}}
\def\mD{\mathcal{D}}
\def\mE{\mathcal{E}}
\def\mF{\mathcal{F}}
\def\mG{\mathcal{G}}
\def\mH{\mathcal{H}}
\def\mI{\mathcal{I}}
\def\mJ{\mathcal{J}}
\def\mK{\mathcal{K}}
\def\mL{\mathcal{L}}
\def\mM{\mathcal{M}}
\def\mN{\mathcal{N}}
\def\mO{\mathcal{O}}
\def\mP{\mathcal{P}}
\def\mQ{\mathcal{Q}}
\def\mR{\mathcal{R}}
\def\mS{\mathcal{S}}
\def\mT{\mathcal{T}}
\def\mU{\mathcal{U}}
\def\mV{\mathcal{V}}
\def\mW{\mathcal{W}}
\def\mX{\mathcal{X}}
\def\mY{\mathcal{Y}}
\def\mZ{\mathcal{Z}}

\def\1n{\mathbf{1}_n}
\def\0{\mathbf{0}}
\def\1{\mathbf{1}}

\def\A{{\bf A}}
\def\B{{\bf B}}
\def\C{{\bf C}}
\def\D{{\bf D}}
\def\E{{\bf E}}
\def\F{{\bf F}}
\def\G{{\bf G}}
\def\H{{\bf H}}
\def\I{{\bf I}}
\def\J{{\bf J}}
\def\K{{\bf K}}
\def\L{{\bf L}}
\def\M{{\bf M}}
\def\N{{\bf N}}
\def\O{{\bf O}}
\def\P{{\bf P}}
\def\Q{{\bf Q}}
\def\R{{\bf R}}
\def\S{{\bf S}}
\def\T{{\bf T}}
\def\U{{\bf U}}
\def\V{{\bf V}}
\def\W{{\bf W}}
\def\X{{\bf X}}
\def\Y{{\bf Y}}
\def\Z{{\bf Z}}

\def\a{{\bf a}}
\def\b{{\bf b}}
\def\c{{\bf c}}
\def\d{{\bf d}}
\def\e{{\bf e}}
\def\f{{\bf f}}
\def\g{{\bf g}}
\def\h{{\bf h}}
\def\i{{\bf i}}
\def\j{{\bf j}}
\def\k{{\bf k}}
\def\l{{\bf l}}
\def\m{{\bf m}}
\def\n{{\bf n}}
\def\o{{\bf o}}
\def\p{{\bf p}}
\def\q{{\bf q}}
\def\r{{\bf r}}
\def\s{{\bf s}}
\def\t{{\bf t}}
\def\u{{\bf u}}
\def\v{{\bf v}}
\def\w{{\bf w}}
\def\x{{\bf x}}
\def\y{{\bf y}}
\def\z{{\bf z}}

\def\balpha{\mbox{\boldmath{$\alpha$}}}
\def\bbeta{\mbox{\boldmath{$\beta$}}}
\def\bdelta{\mbox{\boldmath{$\delta$}}}
\def\bgamma{\mbox{\boldmath{$\gamma$}}}
\def\blambda{\mbox{\boldmath{$\lambda$}}}
\def\bsigma{\mbox{\boldmath{$\sigma$}}}
\def\btheta{\mbox{\boldmath{$\theta$}}}
\def\bomega{\mbox{\boldmath{$\omega$}}}
\def\bxi{\mbox{\boldmath{$\xi$}}}
\def\bnu{\mbox{\boldmath{$\nu$}}}                                  
\def\bphi{\mbox{\boldmath{$\phi$}}}
\def\bmu{\mbox{\boldmath{$\mu$}}}

\def\bDelta{\mbox{\boldmath{$\Delta$}}}
\def\bOmega{\mbox{\boldmath{$\Omega$}}}
\def\bPhi{\mbox{\boldmath{$\Phi$}}}
\def\bLambda{\mbox{\boldmath{$\Lambda$}}}
\def\bSigma{\mbox{\boldmath{$\Sigma$}}}
\def\bGamma{\mbox{\boldmath{$\Gamma$}}}

\newcommand{\myminimum}[1]{\mathop{\textrm{minimum}}_{#1}}
\newcommand{\mymaximum}[1]{\mathop{\textrm{maximum}}_{#1}}    
\newcommand{\mymin}[1]{\mathop{\textrm{minimize}}_{#1}}
\newcommand{\mymax}[1]{\mathop{\textrm{maximize}}_{#1}}
\newcommand{\mymins}[1]{\mathop{\textrm{min.}}_{#1}}
\newcommand{\mymaxs}[1]{\mathop{\textrm{max.}}_{#1}}  
\newcommand{\myargmin}[1]{\mathop{\textrm{argmin}}_{#1}} 
\newcommand{\myargmax}[1]{\mathop{\textrm{argmax}}_{#1}} 
\newcommand{\myst}{\textrm{s.t. }}

\newcommand{\denselist}{\itemsep -1pt}
\newcommand{\sparselist}{\itemsep 1pt}

\definecolor{pink}{rgb}{0.9,0.5,0.5}
\definecolor{purple}{rgb}{0.5, 0.4, 0.8}   
\definecolor{gray}{rgb}{0.3, 0.3, 0.3}
\definecolor{mygreen}{rgb}{0.2, 0.6, 0.2}

\newcommand{\cyan}[1]{\textcolor{cyan}{#1}}
\newcommand{\red}[1]{\textcolor{red}{#1}}  
\newcommand{\blue}[1]{\textcolor{blue}{#1}}
\newcommand{\magenta}[1]{\textcolor{magenta}{#1}}
\newcommand{\pink}[1]{\textcolor{pink}{#1}}
\newcommand{\green}[1]{\textcolor{green}{#1}} 
\newcommand{\gray}[1]{\textcolor{gray}{#1}}    
\newcommand{\mygreen}[1]{\textcolor{mygreen}{#1}}    
\newcommand{\purple}[1]{\textcolor{purple}{#1}}       

\definecolor{greena}{rgb}{0.4, 0.5, 0.1}
\newcommand{\greena}[1]{\textcolor{greena}{#1}}

\definecolor{bluea}{rgb}{0, 0.4, 0.6}
\newcommand{\bluea}[1]{\textcolor{bluea}{#1}}
\definecolor{reda}{rgb}{0.6, 0.2, 0.1}
\newcommand{\reda}[1]{\textcolor{reda}{#1}}

\def\changemargin#1#2{\list{}{\rightmargin#2\leftmargin#1}\item[]}
\let\endchangemargin=\endlist
                                               
\newcommand{\cm}[1]{}

\newcommand{\mtodo}[1]{{\color{red}$\blacksquare$\textbf{[TODO: #1]}}}
\newcommand{\myheading}[1]{\vspace{1ex}\noindent \textbf{#1}}
\newcommand{\htimesw}[2]{\mbox{$#1$$\times$$#2$}}
\newcommand{\mh}[1]{\textcolor{magenta}{
\ifdraft [Minh: {#1}]
\fi}}
\newcommand{\ms}[1]{\textcolor{red}{[MS: {#1}]}}

\newif\ifshowsolution
\showsolutiontrue

\ifshowsolution  
\newcommand{\Comment}[1]{\paragraph{\bf $\bigstar $ COMMENT:} {\sf #1} \bigskip}
\newcommand{\Solution}[2]{\paragraph{\bf $\bigstar $ SOLUTION:} {\sf #2} }
\newcommand{\Mistake}[2]{\paragraph{\bf $\blacksquare$ COMMON MISTAKE #1:} {\sf #2} \bigskip}
\else
\newcommand{\Solution}[2]{\vspace{#1}}
\fi

\newcommand{\truefalse}{
\begin{enumerate}
	\item True
	\item False
\end{enumerate}
}

\newcommand{\yesno}{
\begin{enumerate}
	\item Yes
	\item No
\end{enumerate}
}
\newcommand{\Sref}[1]{Sec.~\ref{#1}}
\newcommand{\Eref}[1]{Eq.~(\ref{#1})}
\newcommand{\Fref}[1]{Fig.~\ref{#1}}
\newcommand{\Tref}[1]{Tab.~\ref{#1}}


%% file: sec-intro.tex
Localization of people or objects, i.e., identifying the location of each instance, in a crowded scene is an important problem for many fields. 
Localization of people, animals, or biological cells provides detailed spatial information that can be crucial in journalism~\cite{counting:journalism:contexts:2004}, ecology \cite{counting:ecology:2008} or cancer research \cite{counting:cancer:2018}. 
A high quality localization algorithm naturally solves the popular crowd counting problem, i.e., counting the number of people in a crowded scene \cite{qnrf:Idrees:CompositionLF:eccv:2018}. 
Furthermore, the rich spatial pattern can be used in many other tasks, e.g., initialization of tracking algorithms \cite{tracking:fuse:cvpr:2018}, animal population studies \cite{counting:ecology:2008}, tumor microenvironment analyses \cite{chao:micro:socg:2020}, and monitoring of social distancing \cite{yang2020vision}.

\begin{figure*}
\centering
\includegraphics[width=1\linewidth]{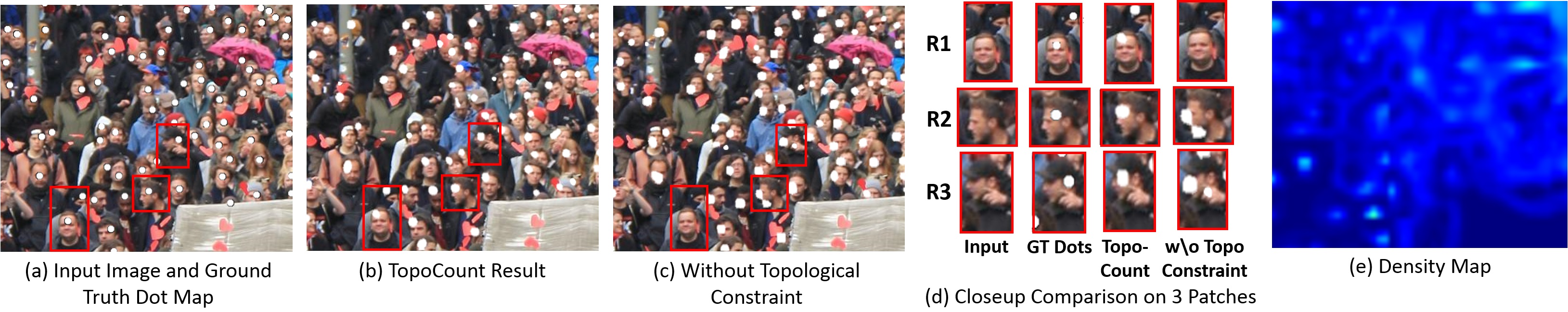}
\caption{(a) a sample image and the ground truth (GT) dots; (b) the localization result of our method (TopoCount); (c) without the topological constraints, topological errors may happen and the localization quality is impacted; (d) closeup view of some specific patches. Without the topological constraints, the prediction often misses dots or collapses nearby dots (R1). It can also create phantom dots (R2 and R3). TopoCount successfully avoids such errors; (e) a density map from a SOTA counting method \cite{bayesian:2019:ICCV}. The density map loses important topological information and cannot recover spatial arrangement of dots.} 
\label{fig:anno}
\end{figure*}

Despite many proposed methods \cite{detection2:tpami:2008,detection3:2009:cvpr,RAZ-Net:detect:cvpr:2019,locate:LSC-CNN:babu:ipmi:2020}, localization remains a challenging task. Aside from fundamental challenges of a crowded scene such as perspective, occlusion, and cluttering, one key issue is the limitation of annotation. Due to the large number of target instances, the ground truth annotation is usually provided in the form of dots located inside the instances (Fig.~\ref{fig:anno}(a)). These dots only provide limited information. 
A dot can be arbitrarily perturbed as long as it is within the target instance, which can be of very different scales.  
As a consequence, the dot features are not specific.
Without sufficient supervision, we cannot decide the boundary between instances. Thus, it is very hard to prevent spatial semantic errors, i.e., 
predicting multiple dots within a same person (false positives) or collapsing the dots of multiple persons in a cluttered area (false negatives). 

In this paper, we propose a novel topological approach for the localization problem. 
We treat the problem as predicting a binary mask, called the Topological Map (Fig.~\ref{fig:anno}(b)), whose connected components one-to-one correspond to the target dots. 
The number of components in the predicted mask should be the same as the number of ground truth dots. This spatial semantic constraint is indeed \emph{topological}. 
During training we enforce such a ``topological constraint'' locally, i.e., the topology should be correct within each randomly sampled patch. 
The topological constraint teaches the model to reason about spatial arrangement of dots and avoids incorrect phantom dots and collapsing dots. 
This significantly improves the  localization method quality, especially near dense regions. 
 See Fig.~\ref{fig:anno}(b), (c) and (d).

To enforce the topological constraint, we introduce a novel loss, called \emph{persistence loss}, based on the theory of persistent homology \cite{edelsbrunner2010computational}.
Instead of directly computing the topology of the predicted binary mask, we inspect the underlying likelihood map, i.e., the sigmoid layer output of the neural network. The   persistent homology algorithm captures the topographic landscape features of the likelihood map, namely, modes and their saliency.  
Our persistence loss compares these modes and the true topology. 
 Within a sample patch, if there are $k$ true dots, the persistence loss promotes the saliency of the top~ $k$ modes and penalizes the saliency of the remaining modes.
This way it ensures that there are exactly $k$ modes in the likelihood landscape, all of which are salient. A 0.5-thresholding of such a topologically correct likelihood map gives a binary mask with exactly $k$ components, as desired.

We evaluate our method on various benchmarks and show that our proposed method, TopoCount, outperforms previous localization methods in various localization metrics.

~

\myparagraph{Application to crowd counting.} We further demonstrate the power of our localization method by applying it to a closely related problem, \emph{crowd counting}. For the counting problem, training images are also annotated with dots, but the task is simpler; one only needs to predict the total number of instances in each image. 
State-of-the-art (SOTA) counting algorithms, such as  
 \cite{can:cvpr:2019,bayesian:2019:ICCV, attention-scaling:jiang:2020:CVPR,wang2020DMCount}, 
learn to approximate a density map of the crowd whose integral gives the total count in the image. 
The learnt density function, even with accurate counting number, can significantly lose the topological characterization of the target density, especially near the dense population region. 
See Fig.~\ref{fig:anno}(e).

To solve the counting problem, we can directly apply the localization algorithm and count the number of dots in the output map. However, this is not necessarily ideal for the task. Counting is an easier problem than localization. It has been shown that relaxing the output to a density function is the most effective strategy, although it will lose the exact locations of people or objects.

To achieve the best counting performance, we incorporate our localization result as complimentary information for density-based counting algorithms. 
By introducing our TopoCount results as additional input to SOTA counting algorithms \cite{bayesian:2019:ICCV,can:cvpr:2019}, we improve their counting performance by 7 to 28 \% on several public benchmarks. 
This further demonstrates the power of the spatial configuration information that we obtain through the topological reasoning.

\myparagraph{In summary}, our technical contribution is three-fold.
\begin{itemize}[topsep=0pt,itemsep=0pt,partopsep=0pt, parsep=0pt]
\item We propose a topological constraint to address the topological errors in crowd localization.
\item To enforce the constraint, we propose a novel persistence loss based on the theory of persistent homology.
Our method achieves SOTA localization performance.
\item We further integrate the topology-constrained localization algorithm into density-based counting algorithms to improve the performance of SOTA counting methods.
\end{itemize}

%% file: sec-relatedwork.tex
We discuss various localization approaches; some of which learn the localization algorithm jointly with the counting model.  \citet{locate:LSC-CNN:babu:ipmi:2020} learn to predict bounding boxes of human heads by fusing multi-scale features. 
The model is trained with cross entropy loss over the whole image and special focus on selected high error regions.
\citet{decidenet:dmap:detect:2018:cvpr} performs detection using Faster RCNN \cite{fasterrcnn:nips:2015}. Faster RCNN has been shown to not scale well with the increasing occlusion and clutter in crowd counting benchmarks \cite{gao:nwpu:tpmi:2020}.
 \citet{RAZ-Net:detect:cvpr:2019} also learn to predict a localization map as a binary mask. They use a weighted cross-entropy loss to compensate for the unbalanced foreground/background pixel populations. In dense regions, the localization is further improved by recurrent zooming. 
However, all these methods are not explicitly modeling the topology of dots and thus cannot avoid topological errors (phantom dots and dots collapsing).
 
A related method is by \cite{blobs:detect:eccv:2018}. It formulates the problem as a semantic segmentation problem. Blobs of the segmentation mask correspond to the target object instances. A blob is split if it contains multiple true dots and is suppressed if it does not contain any true dot. This method is not robust to the perturbation of dot locations; a blob that barely misses its corresponding true dot will be completely suppressed. On the contrary, our method leverages the deformation-invariance of topological structures arising from dots, and thus can handle the dot perturbation robustly.

\cite{iterativeCC:Ranjan:eccv:2018,sanet:eccv:2018,csr:cvpr:2018,can:cvpr:2019,bayesian:2019:ICCV,attention-scaling:jiang:2020:CVPR,m_Ranjan-etal-ACCV20, wang2020DMCount}. 
These methods train a neural network to generate a density function, the integral of which represents the estimated object count \cite{oxford:nips:2010}. The ground truth density functions are generated by Gaussian kernels centered at the dot locations. While these methods excel at counting, the smoothed density functions lose the detailed topological information of the original dots, especially in dense areas (see Fig.~\ref{fig:anno}(e)). As a consequence, localization maps derived from the estimated density maps, e.g., via integer programming \cite{smallobj:2015:cvpr} or via multi-scale representation of the density function \cite{qnrf:Idrees:CompositionLF:eccv:2018}, are also of limited quality.

Topological information has been used in various learning and vision tasks. Examples include but are not limited to shape analysis \cite{reininghaus2015stable,carriere2017sliced}, graph learning \cite{hofer2017deep,zhao2019learning,zhao2020persistence}, clustering \cite{ni2017composing,chazal2013persistence}, learning with label noise \cite{wu2020topological} and image segmentation \cite{wu2017optimal,mosinska:topo:delineation:cvpr:2018,chan:topo:beltrami:deformable:2017,waggoner:topo:multilabel:wccv:2015}. 
Persistent-homology-based objective functions have been used for image segmentation \cite{xiaoling:topo:segmentation:nips:2019,clough:topo:priorseg:ipmi:2019}, generative adversarial networks (GANs) \cite{wang2020topogan}, graphics \cite{poulenard2018topological} and machine learning model regularization \cite{hofer2019connectivity,chen2019topological}.
To the best of our knowledge, our method is the first to exploit topological information in crowd localization and counting tasks, and to use a topology-informed loss to solve the corresponding topological constraint problem.

%% file: sec-method.tex
\begin{figure}[b!]
\centering
\includegraphics[width=1.0\linewidth]{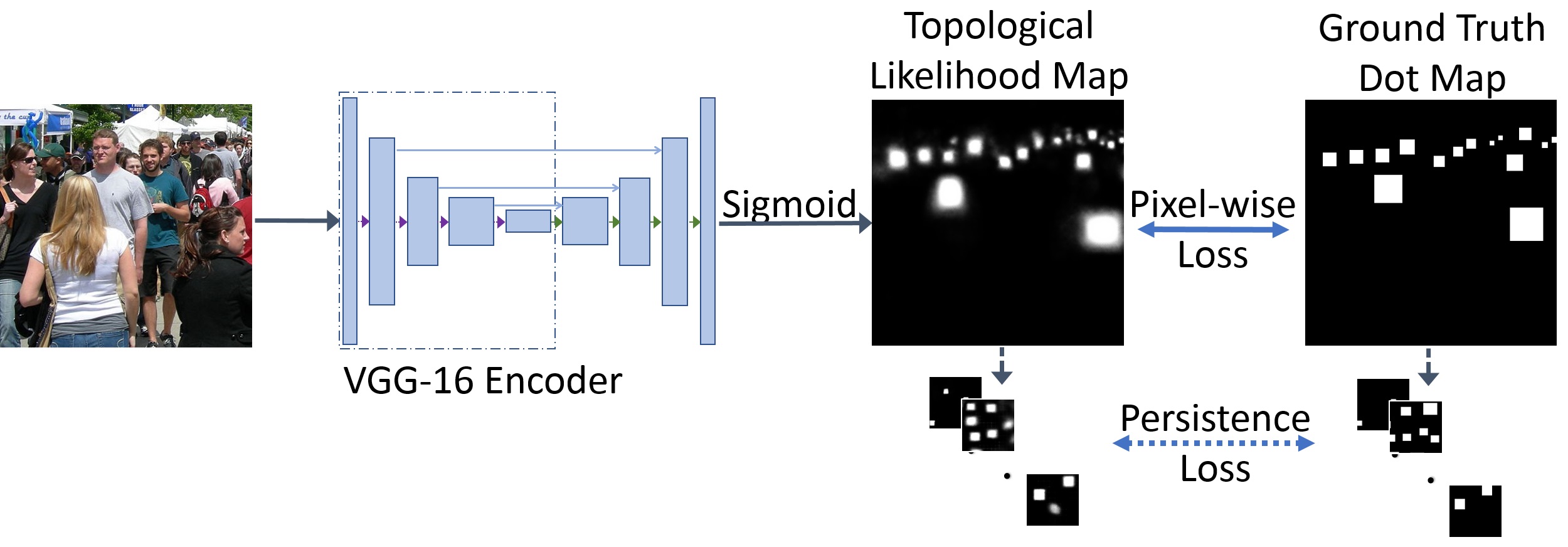}
\caption{TopoCount has a U-Net style architecture with a VGG-16 encoder. During training a pixel-wise loss (DICE loss) is applied on the whole image and persistence loss is applied on sampled patches. 
}
\label{fig:arch}
\end{figure}

\begin{figure*}
\centering
\includegraphics[width=1.0\linewidth]{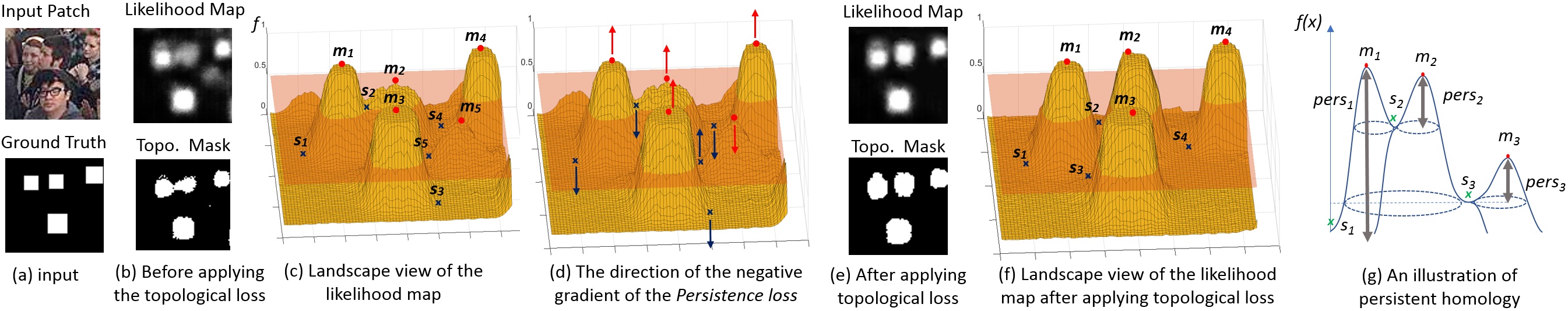}
\caption{
(a) An example image patch and its ground truth dot map (with 4 true dots). 
(b) The likelihood map and prediction mask without topological constraint. 
(c) A landscape view of the likelihood function. There are 5 modes (red dots) and their paired saddles(blue crosses). The top 4 salient ones ($m_1$ to $m_4$) are matched to the ground truth dots. The 5th, $m_5$, is not matched. The thresholding excludes the weak mode ($m_5$) in the predicted mask. But $m_1$ and $m_2$ are merged in the thresholded result because the saddle point between them ($s_2$) is above the cutoff value.
(d) Optimizing the Persistence loss will suppress $m_5$ by reducing $f(m_5)$. 
Meanwhile, it will enhance the saliency of $m_2$ by increasing $f(m_2)$ and decreasing $f(s_2)$. When $f(s_2)$ is below the threshold, $m_1$ and $m_2$ are separated into two components in the final prediction.
(e) The likelihood and the prediction mask with topological constraint (persistence loss). 
The collapsing of $m_1$ and $m_2$ is avoided.
(f) A landscape view of the likelihood function in (e). Only 4 modes remain.
(g) An illustration of persistent homology. Three modes are paired with saddles at which their attractive basins merge with others. The differences $f(m_i) - f(s_i)$, $i = 1, 2,3$ are their persistence.
}
\label{fig:persloss}
\end{figure*}
We formulate the localization problem as a structured prediction problem. Given training images labeled with  \emph{dot annotations}, i.e., sets of dots representing persons (Fig.~\ref{fig:anno}(a)), 
we train our model to predict a binary mask. Each connected component in the mask represents one person. We take the centers of the connected components as the predicted dots. For training, we expand the dot annotations of training images into dot masks by a slight dilation of each dot, but with the condition that the expanded dots do not overlap. 
We call this ``dot mask'' the \emph{ground truth dot map}.

To train a model to predict this binary ground truth dot map, we use a U-Net type architecture with a per-pixel loss. 
The output of the model after the Sigmoid activation is called the \emph{topological likelihood map}. During inference, a final thresholding step is applied to the likelihood to generate the binary mask. We call the mask the \emph{topological dot map} as it is required to have the same topology as the ground truth dot map.
Fig.~\ref{fig:arch} shows our overall architecture.

The rest of this section is organized as follows. We first introduce the topological constraint for the topological dot map. Next, we formalize the persistence loss that is used to enforce the topological constraint. Afterwards, we provide details of the architecture and training.
Finally, we discuss how to incorporate our method into SOTA counting algorithms to improve their counting performance.
\subsection{Topological Constraint for Localization}
\label{sec:topo_motive}
For the localization problem, a major challenge is the perturbation of dot annotation. In the training dot annotation, a dot can be at an arbitrary location of a person and can correspond to different parts of a human body. Therefore it is hard to control the spatial arrangement of the predicted dots. As illustrated in Fig.~\ref{fig:anno}(d), a model without special design can easily predict multiple ``phantom dots'' at different body parts of the same person. At cluttered regions, the model can exhibit ``dots collapsing''. 
To address these semantic errors, we must teach the model to learn the spatial contextual information and to reason about the interactions between dots. A model needs to know that nearby dots are mutually exclusive if there are no clear boundary between them. It should also encourage more dots at cluttered regions.
To teach the model this spatial reasoning of dots, we define a \emph{topological constraint} for the predicted topological dot map $y$: 


\begin{definition}[Topological constraint for localization] Within any local patch of size $h\times w$, the Betti number of dimension zero; i.e, the number of connected components, of $y$ equals to the number of ground truth dots. 
\end{definition}

\smallskip
This constraint allows us to encode the spatial arrangement of dots effectively without being too specific about their locations. This way the model can avoid the topological errors such as phantom dots and dots collapsing, while being robust to perturbation of dot annotation. Next, we introduce a novel training loss to enforce this topological constraint.
\subsection{Persistence Loss}
Directly enforcing the topological constraint in training is challenging. The number of connected components and the number of dots within each patch are discrete values and their difference is non-differentiable. We introduce a novel differentiable loss called \emph{persistence loss}, based on the persistence homology theory \cite{edelsbrunner2000topological,edelsbrunner2010computational}. 
The key idea is that instead of inspecting the topology of the binary topological dot map, we use the continuous-valued likelihood map of the network $f$. 
We consider $f$ as a terrain function and consider the landscape features of the terrain. These features provide important structural information. In particular, we focus on the modes (i.e., local maxima) of $f$. As illustrated in Fig.~\ref{fig:persloss}(b)(c), a salient mode of $f$, after thresholding, will become a connected component in the predicted topological dot map. A weak mode will miss the cutoff value and disappear in the dot map. 

The persistence loss captures the saliency of modes and decides to enhance/suppress these modes depending on the ground truth topology.
Given a patch with $c$ many ground truth dots, our persistence loss enforces the likelihood $f$ to only have $c$ many salient modes, and thus $c$ connected components in $y$. It reinforces the total saliency of the top $c$ modes of $f$, and suppresses the saliency of the rest. The saliency of each mode, $m$, is measured by its \emph{persistence}, $\pers(m)$, which will be defined shortly. As an example, in Fig.~\ref{fig:persloss}(c), $f$ has $5$ salient modes. If $c=4$, the persistence loss will suppress the mode with the least persistence, in this case $m_5$, and will reinforce the other 4 modes. As a consequence, the mode $m_2$ is enhanced and is separated from $m_1$, avoiding a mode collapsing issue. Formally: 
\begin{definition}[Persistence Loss] Given a patch, $\delta$, with $c$ ground truth dots, denote by $\calM_{c}$ the top $c$ salient modes, and $\overline{\calM_{c}}$ the remaining modes of $f$. The persistence loss of $f$ at the patch $\delta$ is
\begin{equation}
\label{eq:persloss}
    \Lpers(f,\delta) = - \sum\nolimits_{m\in \calM_{c}} \pers(m) + \sum\nolimits_{m\in \overline{\calM_{c}}} \pers(m)
    \end{equation}
\end{definition}
Minimizing this loss is equivalent to maximizing the saliency of the top $c$ modes and minimizing the saliency of the rest. 
Consequently, the function will only have $c$ salient modes, corresponding to $c$ components in the predicted mask,  
Fig.~\ref{fig:persloss}(e)(f).
Next we formalize the mode saliency, called  \emph{persistence}, and derive the gradient of the loss.

\myparagraph{Saliency/persistence of a mode.}
For a mode $m$ (local maximum), its \emph{basin of attraction} is the region of all points from which a gradient ascent will converge to $m$.
Intuitively, 
the persistence of $m$, measuring its “relative height”, is the difference
between its height $f(m)$ and the level $f(s)$ at which its basin of attraction meets that of another higher mode. See Fig.~\ref{fig:persloss}(g) for an illustration.

 In implementation, the saliency/persistence of each mode is computed by capturing its local maximum and corresponding saddle point. To find each mode $m_i$ and its corresponding saddle point $s_i$ where the component of $m_i$ dies, we use a merging tree algorithm  \cite{edelsbrunner2010computational,ni2017composing}.
A pseudo code of the algorithm is outlined in Appendix~\ref{sec:appendix-topology-algorithm}.

\begin{figure*}[h]
\centering
\makebox[\textwidth][c]{\includegraphics[width=1.0\linewidth]{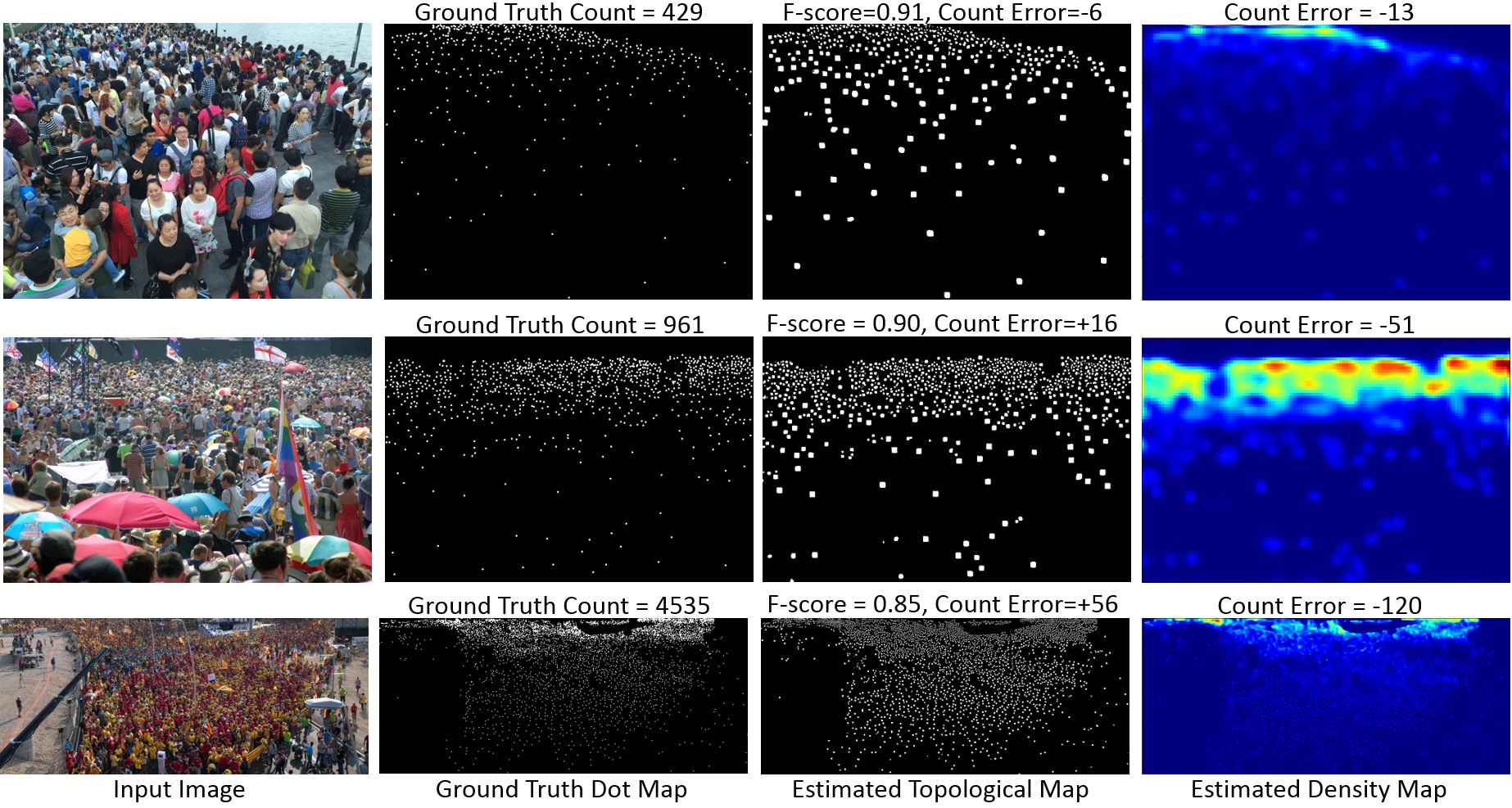}}
\caption{Sample results from different density crowd images. 
The columns represent the original image, ground truth,  topological map by TopoCount, and the estimated density map by the integration of Bayesian + TopoCount. 
}
\label{fig:test_4cc}
\end{figure*}

\newcommand{\calC}{\mathcal{C}}
\newcommand{\calN}{\mathcal{N}}
\newcommand{\calP}{\mathcal{P}}

This algorithm is almost linear. The complexity is $O(n\log n + n \alpha(n))$, where $n$ is the patch size. The $O(n\log n)$ term is due to the sorting of all pixels. $O(n\alpha(n))$ is the complexity for the union-find algorithm for merging connected components. $\alpha(n)$ is the inverse Ackermann's function, which is almost constant in practice. 
The algorithm will detect all critical points, i.e. modes and saddle points, at different thresholds and pair them properly corresponding to all topological features of the function/landscape. 

Having obtained the critical points of the likelihood function using the above algorithm, we apply the persistence loss as follows: For each component $c_i$, denote by $m_i$ its birth maximum and by $s_i$ its death saddle critical points. The persistence of $c_i$ is $\pers(m_i) = f(m_i) - f(s_i)$. 
We sort all modes (or maximum-saddle pairs) according to their persistence. 
The persistence loss in Eq.~\eqref{eq:persloss} can be rewritten as
\begin{eqnarray}
    \Lpers(f,\delta) =&-\sum\nolimits_{m_i \in \calM_c} \left(f(m_i) - f(s_i)\right) \nonumber\\ 
    &+ \sum\nolimits_{m_i \in \overline{\calM_c}} \left(f(m_i) - f(s_i)\right)
\label{eq:persloss2}
\end{eqnarray}

When we take the negative gradient of the loss, for each of the top $c$ modes, we will improve its saliency by increasing the function value at the maximum, $f(m_i)$, and decreasing the function value at its saddle $f(s_i)$. But for each other mode that we intend to suppress, the negative gradient will suppress the maximum's value and increase the saddle point's value. An important assumption in this setting is that the critical points, $m_i$ and $s_i$, are constant when taking the gradient. This is true if we assume a discretized domain and a piecewise linear function~$f$. For this discretized function, within a small neighborhood, the ordering of pixels in function value $f$ remains constant. Therefore the algorithm output of the persistent computation will give the same set of mode-saddle pairs. This ensures that $s_i$ and $m_i$'s for all modes remain constant. The gradient of the loss w.r.t.~the network weights, $W$, 
$\nabla_W\Lpers(f,\delta) = -\sum\nolimits_{m_i \in \calM_c} \left(\frac{\partial f(m_i)}{\partial W} - \frac{\partial f(s_i)}{\partial W}\right) + \sum\nolimits_{m_i \in \overline{\calM_c}} \left(\frac{\partial f(m_i)}{\partial W} - \frac{\partial f(s_i) }{ \partial W}\right)$.

\subsection{TopoCount: Model Architecture and Training}
TopoCount computes the topological map that has the same topology as the dot annotation. To enable the model to learn to predict the dots quickly, we provide per-pixel supervision using DICE loss \cite{dice:2017}.
The DICE loss ($\Ldice$) given Ground truth ($G$) and Estimation ($E$) is: 
$\Ldice(G,E) = 1 - 2 \times \frac{(\sum G \circ E) + 1}{(\sum G^2 + \sum E^2) + 1}$, where  $\circ$ is the Hadamard product.

More precisely, the model is trained with the loss: 
\begin{align}
\mathcal{L} = \Ldice + \lambda_{pers} \Lpers
\label{eqn:total-loss} 
\end{align}
in which $\lambda_{pers}$ adjusts the weight of the persistence loss. An ablation study on the weight $\lambda_{pers}$ is reported in the experiments. 
To provide more balanced samples for the per-pixel loss, we dilate the original dot annotation (treated as a supervision mask) slightly, but ensure that the dilation does not change its topology. The masks of two nearby dots stop dilating if they are about to overlap and impose false topological information.
The size of the dilated dots is not related to the scale of the objects. 
These dilated dot masks from ground truth are used for training. Note that the persistence loss is applied to the likelihood map of the model, $f$.

\myparagraph{Model Architecture Details.}
\label{sec:arch-details}
We use a UNet \cite{unet:2015:miccai} style architecture with a VGG-16 encoder \cite{vgg:Zisserman:2015:ICLR}. The VGG-16 backbone excludes the fully connected layers and has $\approx$ 15 million trainable parameters. There are skip connections between the corresponding encoding and decoding path blocks at all levels except for the first. The skip connection between the first encoder block and last decoder block is pruned to avoid overfitting on low level features, e.g., simple repeated patterns that often occur in crowd areas. 
The final output is the raw topological map, a Sigmoid activation is applied to generate the likelihood map. See Fig~\ref{fig:arch}.
More architecture details are in 
Appendix~\ref{sec:appendix-training-impl-details}.

\subsection{Integration with Counting Methods}
\label{sec:method-integration}
In this section, we discuss how to apply our localization method to the task of crowd counting. In crowd counting, one is given the same dots annotation as the localization task. The goal is to learn to predict the total number of person or objects in an image. A straightforward idea is to use the localization map predicted by TopoCount and directly count the number of dots. Empirically, this solution is already on par with state-of-the-art methods (Table \ref{table:cc_all}). 

Here we present a second solution that is better suited for the counting task. We combine our localization algorithm with existing density-estimation methods \cite{bayesian:2019:ICCV,can:cvpr:2019} to obtain better counting performance. It has been shown that for the counting task, predicting a density map instead of the actual dots is the best strategy, especially in extremely dense or sparse regions. 
We argue that high quality localization maps provide additional spatial configuration information that can further improve the density-estimation counting algorithms.

To combine TopoCount with density-estimation counting methods, 
we use the raw output of a pre-trained TopoCount (the dot map and the topological likelihood map) as two additional channels concatenated to the RGB image. The five-channel `images' are used as the input for a density-estimation model.
The existing density-estimation model has to be adjusted to account for the change in the number of input channels. 
The density estimation model is usually initialized with weights from a pre-trained network on ImageNet.
We keep the architecture and pre-trained weights of the density-estimation model the same everywhere except for the input layer.
We modify the input layer so it accepts five channels and randomly initialize its weights. The density-estimation network is then trained end-to-end.

Our method is agnostic of the density-estimation model.
As a proof-of-concept, we integrate TopoCount with two popular density-estimation counting methods:
Bayesian \cite{bayesian:2019:ICCV} and CAN \cite{can:cvpr:2019}.
We will show that the integration of the localization result learnt with topological constraint significantly boosts the performance of SOTA 
counting algorithms (7 to 28\%, 
see Section~\ref{sec:experiments}). 
This further demonstrates the power of the spatial configuration we obtain through the topological reasoning.

The downside of the combined approach is that it only outputs density maps. The density maps, although better approximates the counts, cannot provide high quality localization information. This is the price one has to pay in order to achieve better counting performance.

%% file: sec-experiments-pre.tex

\setlength{\tabcolsep}{4pt}
\begin{table*}[h]
\small
\begin{center}
\begin{tabular}{|p{0.26\linewidth}|p{0.03\linewidth}|p{0.03\linewidth}|p{0.03\linewidth}|p{0.03\linewidth}|p{0.03\linewidth}|p{0.03\linewidth}|p{0.03\linewidth}|p{0.03\linewidth}|p{0.03\linewidth}|}

\hline
 & \multicolumn{3}{p{0.13\linewidth}|}{\centering {ShanghaiTech A}}  & \multicolumn{3}{p{0.13\linewidth}|}{\centering {ShanghaiTech B}} & \multicolumn{3}{p{0.13\linewidth}|}{\centering {UCF QNRF}}\\
\hline
Model & \centering {G(1)}  & \centering {G(2)}  & \centering {G(3)}  & \centering {G(1)}  & \centering {G(2)}  & \centering {G(3)}  & \centering {G(1)}  & \centering {G(2)}  & \centering {G(3)} \cr 
\hline
CSRNet \cite{csr:cvpr:2018} 
    & \centering 76 & \centering 113 & \centering 149 
    & \centering 13 & \centering 21 & \centering 29 
    & \centering 157 & \centering 187 & \centering 219 \cr 
Bayesian \cite{bayesian:2019:ICCV} 
    & \centering 75 & \centering 90 & \centering 130 
    & \centering \textbf{10} & \centering \textbf{14} & \centering 23 
    & \centering \textbf{100} & \centering \textbf{117} & \centering 150 \cr 
LSC-CNN \cite{locate:LSC-CNN:babu:ipmi:2020} 
    & \centering 70 & \centering 95 & \centering 137 
    & \centering \textbf{10} & \centering 17 & \centering 27 
    & \centering 126 & \centering 160 & \centering 206 \cr 
TopoCount (proposed)  
    & \centering \textbf{69} & \centering \textbf{81} & \centering \textbf{104}
    & \centering \textbf{10} & \centering \textbf{14} & \centering \textbf{20} 
    & \centering 102 & \centering 119  & \centering \textbf{148}\cr 

\hline
\end{tabular}
\caption{Grid Average Mean absolute Errors (GAME) }
\label{table:game}
\end{center}
\vspace{-.1in}
\end{table*}
\setlength{\tabcolsep}{1.4pt}

\begin{table}
\small
\centering
\begin{tabular}{|p{0.53\linewidth}|p{0.13\linewidth}|p{0.13\linewidth}|p{0.13\linewidth}|}
\hline
\centering {Method} & \centering {Prec.}  & \centering {Recall}  & \centering {F-score} \cr 
\hline
MCNN 
    \cite{multicolumn:cc:cvpr:2016} 
    & \centering 59.93\% & \centering 63.50\% & \centering 61.66\% \cr 
CL-CNN D$_{\infty}$ 
    \cite{qnrf:Idrees:CompositionLF:eccv:2018} 
    & \centering 75.8\% & \centering 59.75\% & \centering 66.82\% \cr 
LSC-CNN 
    \cite{locate:LSC-CNN:babu:ipmi:2020} 
    & \centering 74.62\% & \centering 73.50\% & \centering 74.06\% \cr 
TopoCount 
    (proposed)  
    & \centering \textbf{81.77}\% & \centering \textbf{78.96}\% & \centering \textbf{80.34}\% \cr 
\hline
\end{tabular}
\caption{Localization accuracy on the UCF QNRF dataset,  with metric in \cite{qnrf:Idrees:CompositionLF:eccv:2018}}
\label{table:f1score}
\end{table}


\setlength{\tabcolsep}{4pt}
\begin{table}[h]
\small
\begin{center}
\begin{tabular}{|p{0.16\linewidth}|p{0.16\linewidth}|p{0.16\linewidth}|p{0.16\linewidth}|p{0.16\linewidth}|}

\hline
 & \multicolumn{2}{p{0.35\linewidth}|}{\centering {ShanghaiTech A}}  &  \multicolumn{2}{p{0.35\linewidth}|}{\centering {UCF QNRF}}\\
\cline{2-5}
  &  \centering $\sigma=20$  & \centering $\sigma=5$  & \centering $\sigma=20$  & \centering $\sigma=5$   \cr 
  & \centering {mAP/mAR} & \centering {mAP/mAR}   & \centering {mAP/mAR} & \centering {mAP/mAR}   \cr 
\hline
RAZ\_Loc 
    & \centering $58.4/74.1$ & \centering $19.7/42.2$
    & \centering $28.4/48.3$ & \centering $3.7/14.8$
    \cr 
TopoCount 
    & \centering $\textbf{85.0}/\textbf{82.8}$  & \centering $\textbf{56.0}/\textbf{54.8}$
    & \centering $\textbf{69.0}/\textbf{66.5}$  & \centering $\textbf{27.1}/\textbf{26.2}$
    \cr 

\hline
\end{tabular}
\caption{Localization accuracy using metric in \cite{RAZ-Net:detect:cvpr:2019}.}
\label{table:raz_metric1}
\end{center}
\end{table}
\setlength{\tabcolsep}{1.4pt}


\setlength{\tabcolsep}{4pt}
\begin{table}[h]
\small
\begin{center}
\begin{tabular}{|p{0.55\linewidth}|p{0.35\linewidth}|}

\hline
Method  & 
\centering {F1-m / Pre / Rec (\%)}  
\cr
\hline
\multirow{2}{*}{Faster RCNN \cite{fasterrcnn:nips:2015}}
    & \centering $\sigma_l:$ 6.7 / \textbf{95.8} / 3.5  \cr
    & \centering $\sigma_s:$ 6.3 / \textbf{89.4} / 3.3  
     \cr
\hline
\multirow{2}{*}{TinyFaces \cite{tinyfaces:cvpr:2017}} 
    & \centering  $\sigma_l:$ 56.7 / 52.9 / 61.1   \cr
    & \centering  $\sigma_s:$ 52.6 / 49.1 / 56.6  
     \cr
\hline
\multirow{2}{*}{VGG+GPR \cite{vgg:grp:gao2019domainadaptive}} 
    & \centering  $\sigma_l:$ 52.5 / 55.8 / 49.6    \cr
    & \centering  $\sigma_s:$ 42.6 / 45.3 / 40.2  
     \cr
\hline     

\multirow{2}{*}{RAZ\_Loc \cite{RAZ-Net:detect:cvpr:2019} }
    & \centering  $\sigma_l:$ 59.8 / 66.6 / 54.3   \cr
    & \centering $\sigma_s:$ 51.7 / 57.6 / 47.0   
    \cr
\hline    
\multirow{2}{*}{TopoCount (proposed)  }
    & \centering  $\sigma_l:$ \textbf{69.1} / 69.5 / \textbf{68.7}   \cr
    & \centering  $\sigma_s:$ \textbf{60.1} / 60.5 / \textbf{59.8}   
    \cr

\hline
\end{tabular}
\caption{NWPU-Crowd Localization Challenge Results. $\text{F1-m=F1-measure}$. Refer to \cite{gao:nwpu:tpmi:2020} for more details.}
\label{table:nwpu1}
\end{center}
\vspace{-0.1in}
\end{table}
\setlength{\tabcolsep}{1.4pt}

\setlength{\tabcolsep}{4pt}
\begin{table*}[t]
\small
\begin{center}
\begin{tabular}{|p{0.31\linewidth}|p{0.034\linewidth}|p{0.043\linewidth}|p{0.034\linewidth}|p{0.043\linewidth}|p{0.035\linewidth}|p{0.043\linewidth}|p{0.034\linewidth}|p{0.043\linewidth}|p{0.034\linewidth}|p{0.043\linewidth}|p{0.034\linewidth}|p{0.043\linewidth}|}

\hline
 & \multicolumn{2}{p{0.092\linewidth}|}{\centering {Shanghai. A}}  & \multicolumn{2}{p{0.092\linewidth}|}{\centering {Shanghai. B}} & \multicolumn{2}{p{0.093\linewidth}|}{\centering {UCF CC 50}} & \multicolumn{2}{p{0.093\linewidth}|}{\centering {UCF QNRF}} &
 \multicolumn{2}{p{0.092\linewidth}|}{\centering {JHU++}} &
 \multicolumn{2}{p{0.092\linewidth}|}{\centering {NWPU}}\\
\hline
Model 
& \centering {MAE}  & \centering {RMSE}  
& \centering {MAE}  & \centering {RMSE}  
& \centering {MAE}  & \centering {RMSE}  
& \centering {MAE}  & \centering {RMSE}
& \centering {MAE}  & \centering {RMSE}
& \centering {MAE}  & \centering {RMSE}
\cr 
\hline
IC-CNN \cite{iterativeCC:Ranjan:eccv:2018} 
    & \centering 68.5 & \centering 116.2 
    & \centering 10.7 & \centering 16 
    & \centering 260.9 & \centering 365.5 
    & \centering - & \centering -
    & \centering - & \centering -
    & \centering - & \centering - 
    \cr 
CSRNet \cite{csr:cvpr:2018} 
    & \centering 68.2 & \centering 115 
    & \centering 10.6 & \centering 16 
    & \centering 266.1 & \centering 397.5 
    & \centering - & \centering -
    & \centering 85.9  & \centering 309.2
    & \centering 121.3 & \centering 387.8 
    \cr 
SANet \cite{sanet:eccv:2018} 
    & \centering 67 & \centering 104.5 
    & \centering 8.4 & \centering 13.6 
    & \centering 258.4 & \centering 334.9 
    & \centering - & \centering -
    & \centering 91.1  & \centering 320.4
    & \centering 190.6 & \centering 491.4 
    \cr 
ANF \cite{anf:iccv:2019} 
    & \centering 63.9 & \centering 99.4 
    & \centering 8.3 & \centering 13.2 
    & \centering 250.2 & \centering 340 
    & \centering 110 & \centering 174
    & \centering - & \centering -
    & \centering - & \centering - 
    \cr 
RAZ-Net \cite{RAZ-Net:detect:cvpr:2019} 
    & \centering 65.1 & \centering 106.7 
    & \centering 8.4 & \centering 14.1 
    & \centering - & \centering - 
    & \centering 116 & \centering 195
    & \centering - & \centering - 
    & \centering 151.5 & \centering 634.7 
    \cr  
LSC-CNN \cite{locate:LSC-CNN:babu:ipmi:2020} 
    & \centering 66.4 & \centering 117.0 
    & \centering 8.1 & \centering 12.7 
    & \centering 225.6 & \centering 302.7 
    & \centering 120.5 & \centering 218.2
    & \centering 112.7  & \centering 454.4 
    & \centering - & \centering - 
    \cr  
CAN \cite{can:cvpr:2019} 
    & \centering 62.3 & \centering 100 
    & \centering 7.8 & \centering 12.2 
    & \centering 212.2 & \centering \textbf{243.7} 
    & \centering 107 & \centering 183
    & \centering 100.1  & \centering 314.0
    & \centering 106.3 & \centering \textbf{386.5} 
    \cr 
Bayesian \cite{bayesian:2019:ICCV} 
    & \centering 62.8 & \centering 101.8 
    & \centering 7.7 & \centering 12.7 
    & \centering 229.3 & \centering 308.2 
    & \centering {89} & \centering {155}
    & \centering 75.0  & \centering 299.9
    & \centering {105.4} & \centering 454.2 
    \cr 
CG-DRC \cite{sindagi2020jhu-crowd++:dataset} 
    & \centering 60.2 & \centering 94.0 
    & \centering 7.5 & \centering 12.1
    & \centering - & \centering - 
    & \centering 95.5  & \centering 164.3 
    & \centering 71.0  & \centering 278.6 
    & \centering - & \centering - 
    \cr  
ASNet \cite{attention-scaling:jiang:2020:CVPR} 
    & \centering \textbf{57.8} & \centering \textbf{90.1} 
    & \centering - & \centering - 
    & \centering \textbf{174.8}  & \centering 251.6 
    & \centering 91.6  & \centering 159.7
    & \centering - & \centering - 
    & \centering - & \centering - 
    \cr 
DM-Count \cite{wang2020DMCount} 
    & \centering 59.7 & \centering {95.7} 
    & \centering \textbf{7.4} & \centering \textbf{11.8} 
    & \centering 211.0  & \centering 291.5 
    & \centering \textbf{85.6}  & \centering \textbf{148.3}
    & \centering - & \centering - 
    & \centering \textbf{88.4} & \centering 388.6 
    \cr 
\hline

TopoCount (proposed)
    & \centering 61.2 & \centering 104.6 
    & \centering 7.8 & \centering 13.7 
    & \centering 184.1 & \centering 258.3 
    & \centering 89 & \centering 159
    & \centering \textbf{60.9} & \centering \textbf{267.4} 
    & \centering 107.8 & \centering 438.5 
    \cr 

\hline

\hline
\end{tabular}
\caption{Counting Performance Evaluation}
\label{table:cc_all}
\end{center}
\end{table*}
\setlength{\tabcolsep}{1.4pt}


\setlength{\tabcolsep}{4pt}
\begin{table*}[h]
\small
\begin{center}
\begin{tabular}{|p{0.27\linewidth}|p{0.054\linewidth}|p{0.054\linewidth}|p{0.054\linewidth}|p{0.054\linewidth}|p{0.054\linewidth}|p{0.054\linewidth}|p{0.054\linewidth}|p{0.054\linewidth}|p{0.054\linewidth}|p{0.054\linewidth}|}

\hline
 & \multicolumn{2}{p{0.12\linewidth}|}{\centering {ShanghaiTech A}}  & \multicolumn{2}{p{0.12\linewidth}|}{\centering {ShanghaiTech B}} & \multicolumn{2}{p{0.12\linewidth}|}{\centering {UCF CC 50}} & \multicolumn{2}{p{0.12\linewidth}|}{\centering {UCF QNRF}} &
 \multicolumn{2}{p{0.12\linewidth}|}{\centering {JHU++}}\\
\hline
Model & \centering {MAE}  & \centering {RMSE}  & \centering {MAE}  & \centering {RMSE}  & \centering {MAE}  & \centering {RMSE}  & \centering {MAE}  & \centering {RMSE}& \centering {MAE}  & \centering {RMSE}\cr 
\hline

CAN \cite{can:cvpr:2019} 
    & \centering 62.3 & \centering 100 
    & \centering 7.8 & \centering \textbf{12.2} 
    & \centering 212.2 & \centering \textbf{243.7} 
    & \centering 107 & \centering 183
    & \centering 100.1 & \centering 314.0 
    \cr

TopoCount + CAN (proposed)
    & \centering \textbf{59.5 \\ \textit{-4.5\%} }  & \centering \textbf{93.3 \\ \textit{-6.7\%}}  
    & \centering  \textbf{7.5 \\ \textit{-2.5\%}} & \centering 13.2 \\ \textit{+8.1\%}
    & \centering \textbf{190 \\ \textit{-10.5\%}} & \centering 249 \\ \textit{+2.2\%}
    & \centering \textbf{99  \\ \textit{-7.5\%}}& \centering \textbf{162  \\ \textit{-11.5\%}}
    & \centering \textbf{71.9\\ \textit{-28.2\%}} & \centering \textbf{260.9 \\ \textit{-19.9\%}}
    \cr
\hline
Bayesian \cite{bayesian:2019:ICCV} 
    & \centering 62.8 & \centering 101.8 
    & \centering 7.7 & \centering 12.7 
    & \centering 229.3 & \centering 308.2 
    & \centering 89 & \centering 155
    & \centering 75.0 & \centering 299.9 
    \cr 

TopoCount + Bayesian (proposed)
    & \centering \textbf{58 \\ \textit{-7.6\%}}  & \centering \textbf{96.3 \\ \textit{-5.4\%}}  
    & \centering  \textbf{7.2 \\ \textit{-6.5\%}}  & \centering \textbf{11.8 \\ \textit{-7.1\%}} 
    & \centering \textbf{191 \\ \textit{-16.7\%}}  & \centering \textbf{257 \\ \textit{-16.6\%}}  
    & \centering \textbf{85 \\ \textit{-4.5\%}}  & \centering \textbf{148 \\ \textit{-4.5\%}} 
    & \centering \textbf{61.8 \\ \textit{-17.6\%}} & \centering \textbf{262.0 \\ \textit{-12.7\%}} 
    \cr

\hline

\hline
\end{tabular}
\caption{Integration of TopoCount with density estimation methods.}
\label{table:integration}
\end{center}
\end{table*}
\setlength{\tabcolsep}{1.4pt}

\begin{table}[h!]
\small
\begin{center}
\begin{tabular}{|p{0.1\linewidth}|p{0.12\linewidth}|p{0.13\linewidth}|p{0.13\linewidth}|p{0.13\linewidth}|p{0.13\linewidth}|p{0.13\linewidth}|}
\hline

 & \centering BCE & \centering DICE  & \centering $\lambda=0.5$ & \centering $\lambda=1.0$ & \centering $\lambda=1.5$ & \centering $\lambda=2.0$   \cr 
\hline
 {G(3)}  & \centering 122 & \centering 114 & \centering 109 & \centering \textbf{104} & \centering  \textbf{104}
 & \centering 107 \cr 
\hline
\end{tabular}
\caption{Ablation study on the loss function. Compare TopoCount localization score G(3) on the ShanghaiTech Part A dataset when trained with different loss functions: weighted BCE loss, DICE loss ($\lambda=0$), and DICE loss + $\lambda$ Persistence loss ($\lambda = 0.5, 1, 1.5, 2$).} 
\label{table:lambda_pers_vary_sh_a}
\end{center}
\vspace{-.15in}
\end{table}
\setlength{\tabcolsep}{1.4pt}

\begin{table}[htbp]
\small
\begin{center}
\begin{tabular}{|p{0.25\linewidth}|p{0.16\linewidth}|p{0.12\linewidth}|p{0.12\linewidth}|p{0.12\linewidth}|p{0.12\linewidth}|}
\hline
  &&
 \multicolumn{4}{p{0.5\linewidth}|}{\centering {G(L)}} \\ \cline{3-6} 
 \centering Dataset & Patch Size   & \centering G(1) & \centering G(2)  & \centering G(3)  & \centering G(4) 
 \cr 
\hline
\centering Shanghai. A & \centering 150   & \centering 75.4  & \centering 89.9 & \centering 114.2   & \centering - \cr 
\centering & \centering 100   & \centering \textbf{68.4}  & \centering 82.0 & \centering 107.7  & \centering - \cr 
\centering & \centering 50   & \centering 69.3  & \centering \textbf{81.6} & \centering \textbf{104.9}  & \centering - \cr 
\centering & \centering 30   & \centering 75.4  & \centering 86.4 & \centering 108.2  & \centering - \cr 
\hline 
\centering UCF-QNRF & \centering 150   & \centering \textbf{153.5}  & \centering \textbf{175.1} & \centering 208.5  & \centering 272.3   \cr 
\centering & \centering 100   & \centering 155.9  & \centering 177.1 & \centering \textbf{206.6}  & \centering 264.4  \cr 
\centering & \centering 50   & \centering 160.6  & \centering 179.4 & \centering 206.8  & \centering \textbf{260.3}  \cr 
\centering & \centering 30   & \centering 179.9  & \centering 195.2 & \centering 222.1  & \centering 273.1 \cr 
\hline
\end{tabular}
\end{center}
\caption{Comparison of patch size for persistence loss on ShanghaiTech Part A and UCF-QNRF (N=50) using localization score GAME(L)} 
\label{table:patch_pers}
\vspace{-0.1in}
\end{table}
\setlength{\tabcolsep}{1.4pt}

%% file: sec-experiments-post.tex
We validate our method on popular crowd counting benchmarks including ShanghaiTech parts A and B \cite{multicolumn:cc:cvpr:2016}, UCF CC 50 \cite{ucfcc50:idrees:cvpr:2013}, UCF QNRF \cite{qnrf:Idrees:CompositionLF:eccv:2018}, JHU++ \cite{sindagi2020jhu-crowd++:dataset}, and NWPU Challenge \cite{gao:nwpu:tpmi:2020}.

For the localization task, our method is superior compared to other methods. Moreover, we show that the localization results of our method benefits the counting task.

\myparagraph{Training Details.}
We train our \emph{TopoCount} with the dilated ground truth dot mask. The dilation is by default up to 7 pixels. 
For JHU++ and NWPU, which are provided with head box annotation, we use a more accurate dilation guided by the box size, $\max$(7, box width/2, box height/2). In all cases the dilation is no more than half the distance to the nearest neighbor to avoid overlapping of nearby dots. 

 The window size of the patch for topological constraint controls the level of localization we would like to focus on. Since the scale of persons within an image is highly heterogeneous, varying the window size based on scale sounds intriguing. However the ground truth dot annotation generally do not carry scale information. As a result, we fix the patch size for each dataset.
 We use $50{\times}50$ pixels patches for ShanghaiTech and UCF CC 50, and  $100{\times}100$ pixels patches for the larger scale datasets UCF QNRF, JHU++, and NWPU to account for the larger scale variation. 
 An ablation study on the patch size selection is reported in the experiments.
 The persistence loss is applied on grid tiles to enforce topological consistency between corresponding prediction and ground truth tiles/patches.  
 As data augmentation, coordinates of the top left corner of the grid are randomly perturbed. It should be noted that this tiling procedure is only performed during training with the persistence loss and is not performed during inference.

The model is trained with the combined loss $\mathcal{L}$ (Eq.~\eqref{eqn:total-loss}). During the first few epochs the likelihood map is random and is not topologically informative. In the beginning of training we use DICE loss only ($\lambda = 0$). When the model starts to converge to reasonable likelihood maps, we add the persistence loss with  $\lambda = 1$. Fig.~\ref{fig:test_4cc} shows qualitative results. 
More sample results are in 
Appendix~\ref{sec:appendix-experiments}.

\subsection{Localization Performance}
We evaluate \emph{TopoCount} on several datasets using (1) localized counting; (2) F1-score matching accuracy; and (3) the NWPU localization challenge metric.

\myparagraph{Localized Counting.}
We evaluate the counting performance within small grid cells and aggregate the error. The Grid Average Mean absolute Errors (GAME) metric \cite{game:trancos:2015:IbPria}, $G(L)$, divides the image into $4^L$ non-overlapping cells.
In Table~\ref{table:game}, the cell count in the localization-based methods LSC-CNN \cite{locate:LSC-CNN:babu:ipmi:2020} and TopoCount is the sum of predicted dots in the cell. On the other hand, the cell count in the density map estimation methods CSRNet \cite{csr:cvpr:2018} and Bayesian \cite{bayesian:2019:ICCV} is the integral
of the density map over the cell area. TopoCount achieves
the lowest error especially at the finest scale (level L=3), which indicates higher localization accuracy by the predicted dots.

\myparagraph{Matching Accuracy.}
We evaluate matching accuracy in two ways. First, similar to 
\citet{qnrf:Idrees:CompositionLF:eccv:2018} on the UCF QNRF dataset, we perform a greedy matching between detected locations and ground truth dots at thresholds varying from 1 to 100 and average the precision, recall, and F-scores over all thresholds. We compare with scores reported in \cite{qnrf:Idrees:CompositionLF:eccv:2018} in addition to calculated scores for SOTA localization methods \cite{locate:LSC-CNN:babu:ipmi:2020}.  
Table \ref{table:f1score} shows that our method achieves the highest F-score. 

Second, similar to \cite{RAZ-Net:detect:cvpr:2019} on ShanghaiTech Part A and UCF QNRF datasets, we impose at each dot annotation an un-normalized Gaussian function parameterized by $\sigma$. A true positive is a predicted dot whose response to the Gaussian function is greater than a threshold $t$. We compare with \cite{RAZ-Net:detect:cvpr:2019} results at $\sigma=5$ and $\sigma=20$. Table~\ref{table:raz_metric1} reports the mean average precision (mAP) and mean average recall (mAR) for $t \in [0.5, 0.95]$, with a step of $0.05$. TopoCount achieves the highest scores with a large margin at both the small and large sigma $\sigma$.

\myparagraph{NWPU-Crowd Online Localization Challenge.}
NWPU dataset provides dot annotation in addition to box coordinates with specified width $w$ and height $h$ surrounding each head. The online challenge evaluates the F-score with two adaptive matching distance thresholds: $\sigma_l=\sqrt{w^2+h^2}/2$ and a more strict threshold $\sigma_s=min(w, h)/2$. Table \ref{table:nwpu1} shows the F-score, precision, and recall with the 2 thresholds against the published challenge leaderboard. TopoCount achieves the highest F-score in both thresholds. 
More results are in 
Appendix~\ref{sec:appendix-exp-nwpu}.

\subsection{Counting Performance}
Our localization method can be directly applied to crowd counting task. 
It performs competitively among SOTA counting methods.
In Table~\ref{table:cc_all}, we compare TopoCount's overall count in terms of the Mean Absolute Error (MAE) and Root Mean Squared Error (RMSE) against SOTA counting methods. Our method achieves SOTA performance on the new JHU++ large scale dataset and is mostly between second and third place for the other datasets compared to SOTA density-based methods. 
More detailed results on the JHU++ are in 
Appendix~\ref{sec:appendix-exp-jhu}.

\myparagraph{Integration with Density-Based Methods.}
A high quality localization model can be combined with density-based methods to boost their performance. As described in Section~\ref{sec:method-integration}, we integrate TopoCount with two SOTA density-based counting algorithms and report the results.
Table~\ref{table:integration} shows that the integration results in a significant improvement over the individual performance of the density map models. This further demonstrates the high quality of TopoCount localization and suggests that more sophisticated density map models can benefit from high quality localization maps to achieve an even better counting performance.

\subsection{Ablation Studies}

\myparagraph{Ablation Study for the Loss Function.}
On the ShanghaiTech Part A dataset, we compare the performance of TopoCount trained with variations of the loss function in Eq.~\ref{eqn:total-loss}: (1) per-pixel weighted Binary Cross Entropy (BCE) loss as in \cite{RAZ-Net:detect:cvpr:2019} with empirically chosen weight of 5 to account for the amount of class imbalance in the ground truth dot maps, (2) per-pixel DICE loss only (i.e $\lambda=0$ in Eq.~\ref{eqn:total-loss}), and (3) a combined per-pixel DICE loss and Persistence loss with $\lambda \in \{0.5,1,1.5,2\}$. 
The results in Table \ref{table:lambda_pers_vary_sh_a} show the
training with BCE loss gives the largest error. With $\lambda=0$, i.e., DICE without the persistence loss, the error is lower. The error is further lowered with the introduction of the persistence loss. Varying $\lambda$ between $0.5$ and $2.0$ the results are more robust and comparable, with the best performance at $\lambda = 1$ and $\lambda = 1.5$. Consequently, we use $\lambda = 1$ in all our experiments.

\myparagraph{Ablation Study for Choosing Persistence Loss Patch Size}
The window size of the topological constraint patch controls the level of localization we would like to focus on. In the one extreme, when the patch is $1{\times}1$, the topological constraint becomes a per-pixel supervision. It helps the model to learn features for dot pixels,  but loses the rich topological information within local neighborhoods. It is also not flexible/robust with perturbation. On the other extreme, when the patch is the whole image, the topological information is simply the total count of the image. This information is too high-level and will not help the model to learn efficiently; thus we have all other intermediate level supervisions, such as  the density map. A properly chosen patch size will  exploit rich spatial relationships within local neighborhoods while being robust to perturbation. In our experiments, we use a patch size of $50{\times}50$ pixels for ShanghaiTech and UCF CC 50 datasets, for datasets with larger variation in scale, namely UCF QNRF, JHU++, and NWPU-Crowd datasets, we use a larger patch size of $100{\times}100$ pixels. Next we explain how we choose the patch sizes.

To select the  patch size for the persistence loss, we train four models on the ShanghaiTech Part A dataset with different patch sizes: $150{\times}150$, $100{\times}100$, $50{\times}50$, and $30{\times}30$. We evaluate the models localization accuracy using the GAME metric at scales L = 1 through 3, see Table \ref{table:patch_pers}. Training with patch size 30 or 150 yields poor performance. Using patch sizes 50 or 100 gives mostly similar results except at the smallest cell size (L=3) where patch size 50 is the winner, indicating better localization. We thus choose patch size of 50 for the ShanghaiTech and UCF~CC~50 experiments.

The datasets UCF QNRF, JHU++, and NWPU-Crowd are different from the other datasets in their wide variation in scale and resolution. We suspect that a patch size of 50 may not be suitable. We experiment with a small subset of randomly selected (N=50) images from the UCF QNRF training data. Again, we train 4 models with different patch sizes: $150$, $100$, $50$, and $30$, and evaluate the models localization using GAME. Because the images in this dataset have a higher resolution range, we use L = 1 through 4, see Table \ref{table:patch_pers}. We find that a patch size of 150 is more suitable at the coarser cells (L=1, 2) while a  patch size of 50 is more suitable at the finer cells (L=3, 4). For training on these datasets, we choose the intermediate patch size of 100.

%% file: sec-supplemental.tex
\appendix

\section{Persistent Homology Computation}
\label{sec:appendix-topology-algorithm}
\input{appendix-topology-algorithm}

\section{Additional Training and Implementation Details}
\label{sec:appendix-training-impl-details}
\input{appendix-implementation-details}

\section{Additional Experimental Results}
\label{sec:appendix-experiments}
\input{appendix-experiments}

%% file: appendix-topology-algorithm.tex
\begin{algorithm}
\SetAlgoLined
\KwData{Likelihood map $f$}
\KwResult{Paired modes and saddles, $\calP = \{(m_i,s_i)\}$. }
Build a grid graph $G=(V,E)$. Nodes are all pixels. Edges connect adjacent nodes\;
Sort nodes in
decreasing order of their likelihood $f$, $\widehat{V} = (v_1, v_2, \ldots)$, $f(v_i)\geq f(v_{i+1})$\; 
Initialize the visited list $\visited(u)=$ false $\forall u\in V$\;
Initialize a component list, $\calC=\emptyset$\;
Initialize the persistence pair list, $\calP=\emptyset$\;
\ForAll{$u\in \widehat{V}$}{
$\calN_u=\{\text{neighbors of $u$}\}$\;
Components adjacent to $u$: $\calC_u = \emptyset$\;
\tcc{Find all neighbor components of $u$.}
\ForAll{$v\in \calN_u$}{
\If{$\visited(v)$}{
$C_v = $ the component in $\calC$ such that $v\in C_v$\;
$\calC_u = \calC_u \cup \{C_v\}$ 
}

}
\tcc{If $u$ has no neighbor components, it will create a new component.}
\uIf{$|\calC_u|==0$}{  
$C=\{u\}$ 
$\calC = \calC\cup \{C\}$\;
}
\tcc{If $u$ has a single neighbor component, it will merge with it.}
\uElseIf{$|\calC_u|==1$}{
$C = $ the only component in $\calC_u$\;
$C = C\cup \{u\}$\;
}
\tcc{Case $u$ has multiple neighbor components:
(1) Find neighbor component with earliest birth $C_{max}$.
(2) $u$ becomes saddle point for all other neighbor components. 
(3) Merge all other neighbor components with $C_{max}$.
}
\Else{
$C_{max} = \argmax_{C\in \calC_u} \birth(C)$\;
$\calC = (\calC \backslash \calC_u) \cup \{C_{max}\}$\;
\ForAll{$C\in \calC_u\backslash\{C_{max}\}$}{
$m = \argmax_{w\in C} f(w)$\;
$s = u$\;
$\calP = \calP\cup \{(m,s)\}$\;
$C_{max} = C_{max}\cup C$\;
}
}
$\visited(u) = $ true\;
}
\Return $\calP$\;

\caption{Computing Persistence 
}
 \label{alg:pers}
\end{algorithm}

\myparagraph{Persistent homology.}
The theory of persistent homology \cite{edelsbrunner2000topological,edelsbrunner2010computational} measures the saliency of different topological structures from a given scalar function (the likelihood function $f$ in our setting). We threshold an image patch, $\delta$, at a given level $t$ and denote by $\delta_f^t = \{x\in \delta | f(x) \geq t\}$ the \emph{superlevel set}. A superlevel set can have topological structures of different dimensions. For example: 0-dimensional structures are connected components and  1-dimensional structures are handles/holes. 
\footnote{For a more rigorous definition, please refer to classic algebraic topology textbook \cite{munkres2018elements}. }
We focus on 0-dimensional topology in this paper, although the theory covers topology of all dimensions. 
The theory of persistent homology tracks the life span of all topological structures of the superlevel set as we continuously change the threshold $t$.

We
continuously decrease $t$ from $+\infty$ to $-\infty$. As $t$ decreases, we track the connected components of the progressively growing superlevel set, $\delta_f^t$. During the process, a mode (i.e., local maximum)
gives birth to a new connected component. The component dies when it touches another component created by a higher mode. The location at which the two components meet is a saddle point, $s$. The function values of the mode and the saddle point are called the \emph{birth} and \emph{death times}. We use their difference,
called the \emph{persistence}, to measure the saliency of this mode.
See Fig.~2(g) in the paper for an illustration. A pseudo code of the algorithm to find the 0-dimensional components and their persistence is outlined in Algorithm~\ref{alg:pers}.

%% file: appendix-implementation-details.tex
\begin{figure}[htbp]
\centering
\includegraphics[width=1.0\linewidth]{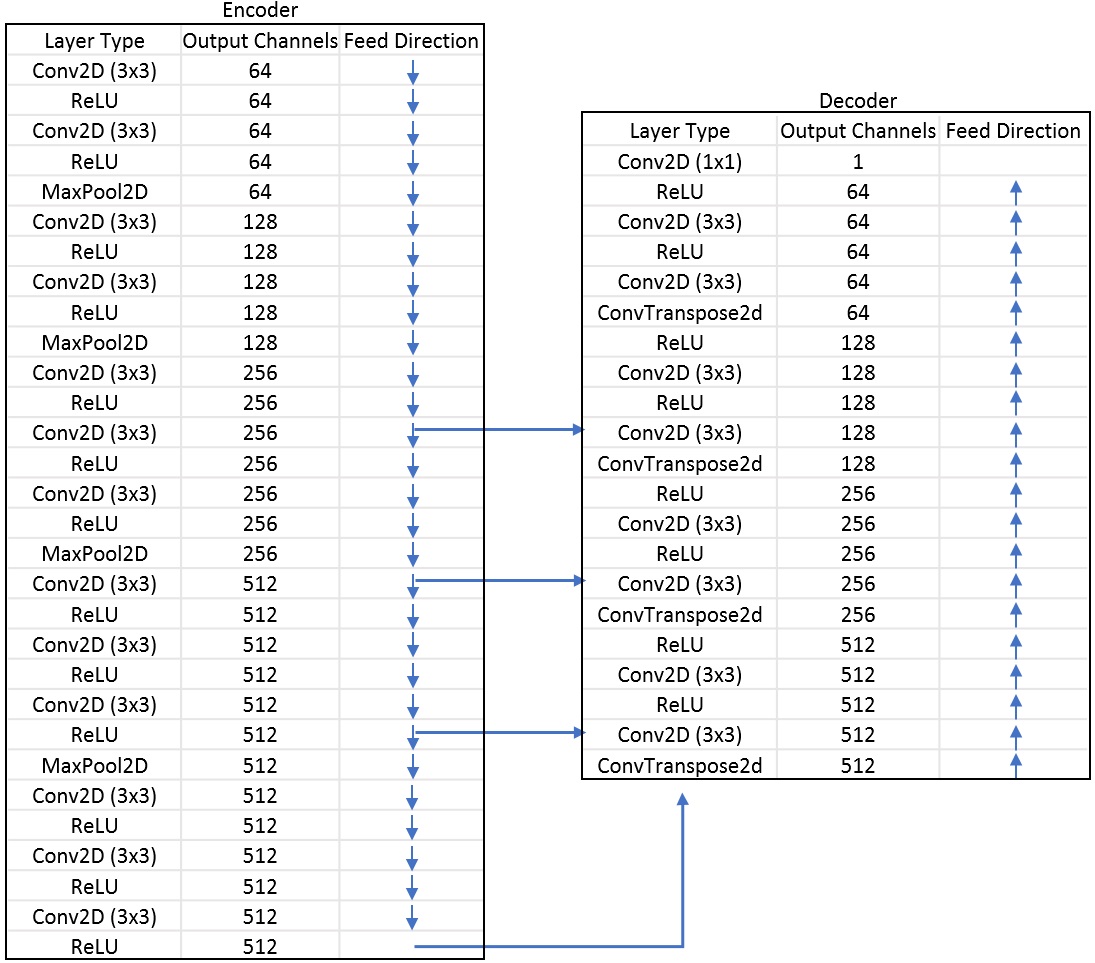}
\caption{TopoCount detailed model architecture. }
\label{fig:arch}
\end{figure}

\myparagraph{Model Architecture}
We use a U-Net style architecture. The detailed per-layer architecture is shown in Fig.~\ref{fig:arch}.

\myparagraph{Image Scaling.}
For all crowd counting datasets we use the original images without scaling, except for the UCF QNRF. For UCF QNRF dataset we apply the following policy: during training the images are resized so that the longer side is of maximum length 2048, and during test we resize so that the shorted side is of maximum length 2048. This more relaxed resizing policy during test allows the model to capture more details in the densely crowded regions of the test images. On the other hand, the variation in scale is not an issue in the JHU++ and NYPU-Crowd datasets because we have extra information of the head box size and the dot dilation size is selected proportionally.

\myparagraph{Training Crop size.}
For the datasets that have an average resolution less than $1024 \times 1024$, i.e. ShanghaiTech Parts A and B, we use the whole image during training. For the rest of the datasets we train with crops of size min(image width,$1024$) $\times$ min(image height,$1024$). 

\myparagraph{Training Optimizer.}
We train TopoCount with Adam optimizer at a learning rate of 0.00005, using a batch size of 1.

\myparagraph{Postprocessing.}
The model generates a likelihood of the topology map. The final mask of the topology map is obtained by thresholding the likelihood. We empirically choose a double thresholding procedure with high threshold = 0.5, and low threshold = 0.4.  In particular, the high threshold is used to first filter the domain and select the connected components representing each person.   Next,  we lower the threshold just to grow the selected connected components.   This is to get the right geometry so that the center of each connected component is closer to the corresponding true dot. The dots are estimated as the centers of the connected components in the mask. The resulting dots are used as the final output and are used in our evaluations with various metrics.

\myparagraph{Software and Hardware}
The implementation used Python 3.6 and Pytorch version 0.5.0a0. The models were training on system with Ubuntu operating system, and NVidia Volta GPU. The amount of GPU memory utilized varies by batch size and crop size. For our configurations it used up less than 10 GB GPU memory.

%% file: appendix-experiments.tex
\subsection{Additional Qualitative Results}

\begin{figure*}[htbp]
\centering
\includegraphics[width=1.0\linewidth]{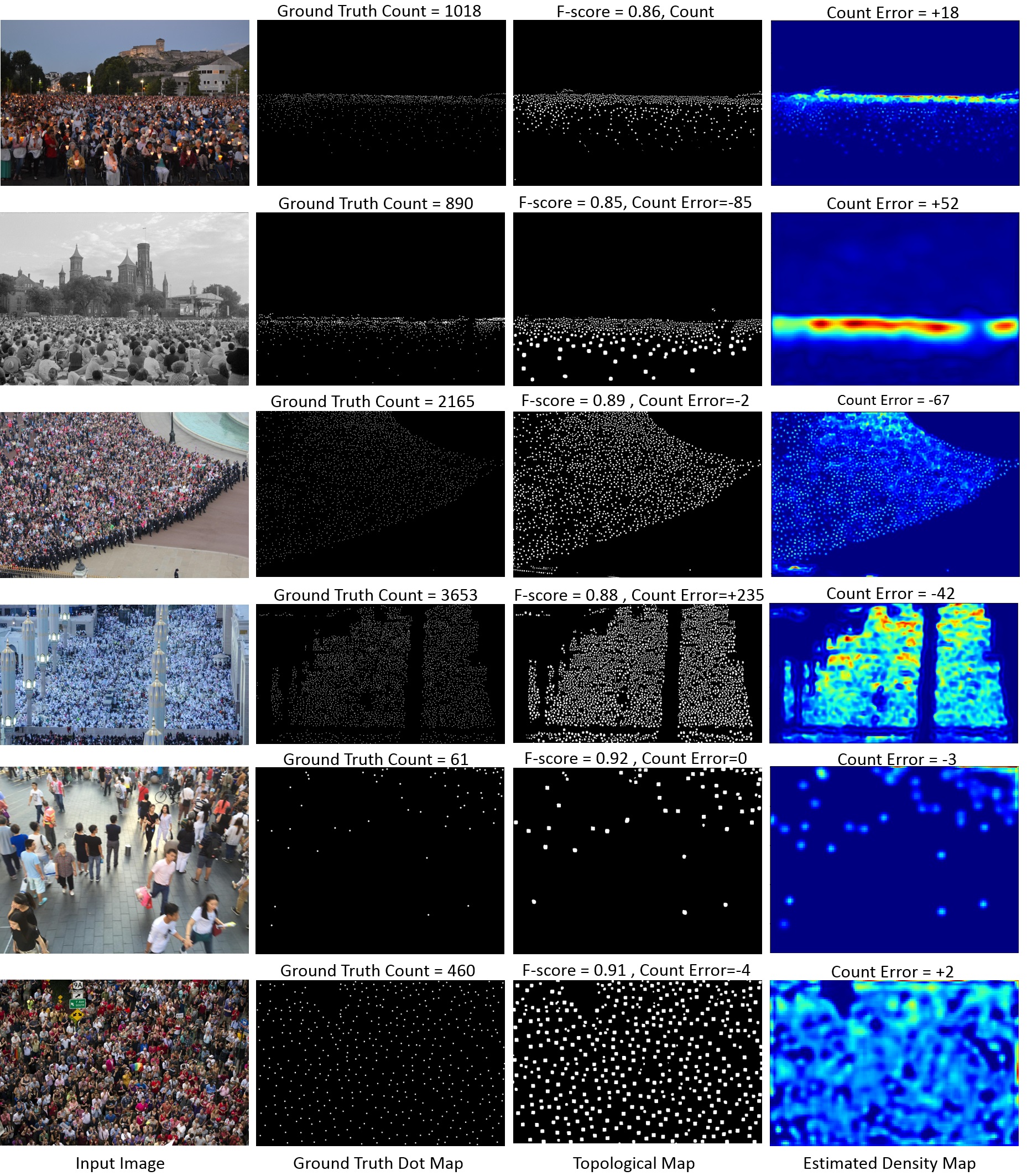}
\caption{
Sample results from different density crowd images. The columns represent the original image, ground truth and topological maps by TopoCount, and the estimated density map by the integration of Bayesian \cite{bayesian:2019:ICCV} + TopoCount.
}
\label{fig:qualitative}
\end{figure*}

Fig.~\ref{fig:qualitative} shows additional qualitative results. It shows samples of the topology and density maps estimated by TopoCount and by Bayesian \cite{bayesian:2019:ICCV} + TopoCount, respectively. The F-scores and counting errors are reported next to the figures. The topological map closely matches the ground truth dots annotation arrangement; as does the density map. Additionally, Fig.~\ref{fig:jhu_a} and Fig.~\ref{fig:jhu_b} show sample results of TopoCount on some difficult cases from the JHU++ dataset. Note that the F-scores reported in all the qualitative results in both the paper and the supplementary material represent the mean of the F-scores calculated using matching distance thresholds ranging from $1$ to $100$, as proposed in \cite{qnrf:Idrees:CompositionLF:eccv:2018}.

\subsection{Integration with Density Map Results}

\begin{figure*}[htbp]
\centering
\includegraphics[width=1.0\linewidth]{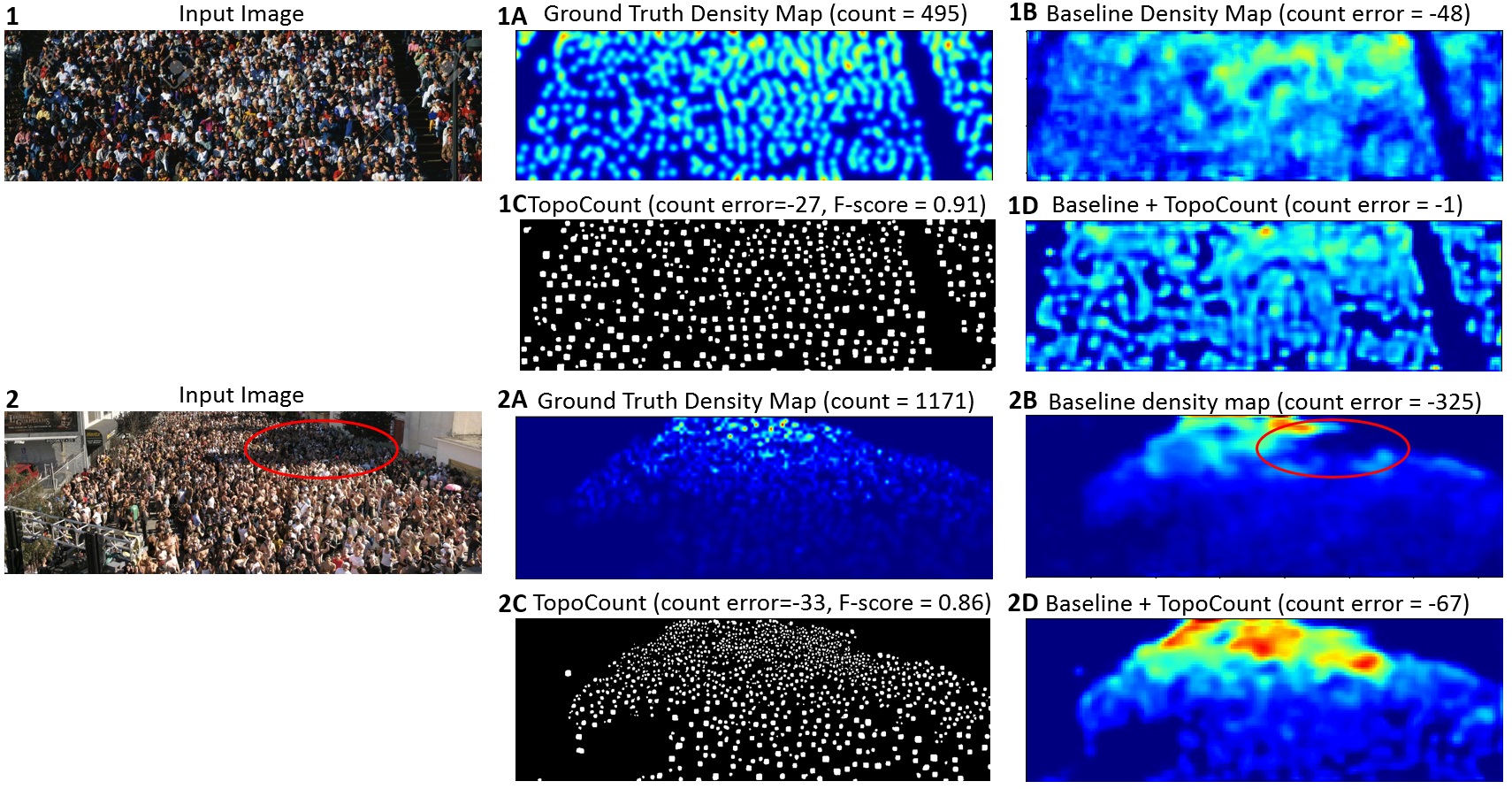}
\caption{Sample results from crowd counting datasets. For each sample we show: the original image, the ground truth density map (A), the baseline density map (by Bayesian loss \cite{bayesian:2019:ICCV}) (B), the topological map by TopoCount (C), and the density map by baseline + TopoCount (D).
With the addition of the topological map, the estimated density map (D) has better topological structure and fixes shadowed regions missed by the baseline \cite{bayesian:2019:ICCV} (B).}
\label{fig:test_dmap}
\end{figure*}

Fig. \ref{fig:test_dmap} shows samples from the density map estimation by the integration of TopoCount and the baseline density map-based method \cite{bayesian:2019:ICCV}. We see in the the figure how the additional information provided by the topological map improves the quality of the density map estimation. In the first sample, the density map of the baseline is blurry while baseline+TopoCount gives a more structured density map that is closer to the ground truth density map. In the second sample, the region indicated by the red ellipse is a densely crowded region that is in the shadow. It is missed by the baseline while TopoCount's topology map is able to identify the crowd. By the integration in baseline+TopoCount, this extra information is passed on to baseline+TopoCount and it also recovers the crowd in the shadow.

\subsection{Shanghai Part A and UCF QNRF Localization Accuracy}
We present more detailed results from the evaluation of the matching accuracy reported in Table \ref{table:raz_metric1} of the paper. We are using the matching metric proposed in \cite{RAZ-Net:detect:cvpr:2019}. At each dot annotation we impose an un-normalized Gaussian function parameterized by $\sigma$. A true positive is a predicted dot whose response to the Gaussian function is greater than a threshold $t$. We compare against \cite{RAZ-Net:detect:cvpr:2019} at $\sigma \in \{40,20,5\}$. The smaller the value of $\sigma$ the closer the prediction needs to be to the ground truth dot to be counted as a true positive. Table~\ref{table:raz_metric2} reports $AP.5$ and $AR.5$: the average precision and recall at $t = 0.5$, and $AP.75$ and $AR.75$: the average precision and recall at $t = 0.75$, in addition to the mean average precision ($mAP$) and mean average recall ($mAR$) for $t \in [0.5, 0.95]$, with a step of $0.05$. TopoCount achieves the highest scores with a large margin at both the small and large sigma $\sigma$.

\subsection{NYPU-Crowd Online Localization Challenge}
\label{sec:appendix-exp-nwpu}
To train on the NWPU-Crowd dataset, we use all training images including those with no heads. 
The model is trained with crops of 1024x1024 pixels of the original image using the original image sizes.
NWPU dataset provides dot annotation in addition to box coordinates with specified width, $w$, and height, $h$, surrounding each head. The online challenge evaluates the F-score with 2 adaptive matching distance thresholds: $\sigma_l=\sqrt{w^2+h^2}/2$ and a more strict threshold $\sigma_s=min(w, h)/2$. 
In Table~\ref{table:nwpu1} of the paper we showed the F-score, precision, and recall with the 2 thresholds against the published challenge leaderboard. 
Here we show more detailed results. For each threshold the recall is further categorized by the head bounding box area range. 
In Table~\ref{table:nwpu2}, $A0 \sim A5$ correspond to the head area ranges: $[10^0, 10^1], (10^1, 10^2], (10^2, 10^3], (10^3, 10^4], (10^4, 10^5]$, and $> 10^5$, respectively.
We see that TopoCount achieves highest recall in the smaller head range catergories while TinyFaces \cite{tinyfaces:cvpr:2017} achieves highest recall in the larger head range catergories.

\subsection{JHU++ Counting Evaluation}
\label{sec:appendix-exp-jhu}
\begin{figure*}[htbp]
\centering
\includegraphics[width=1.0\linewidth]{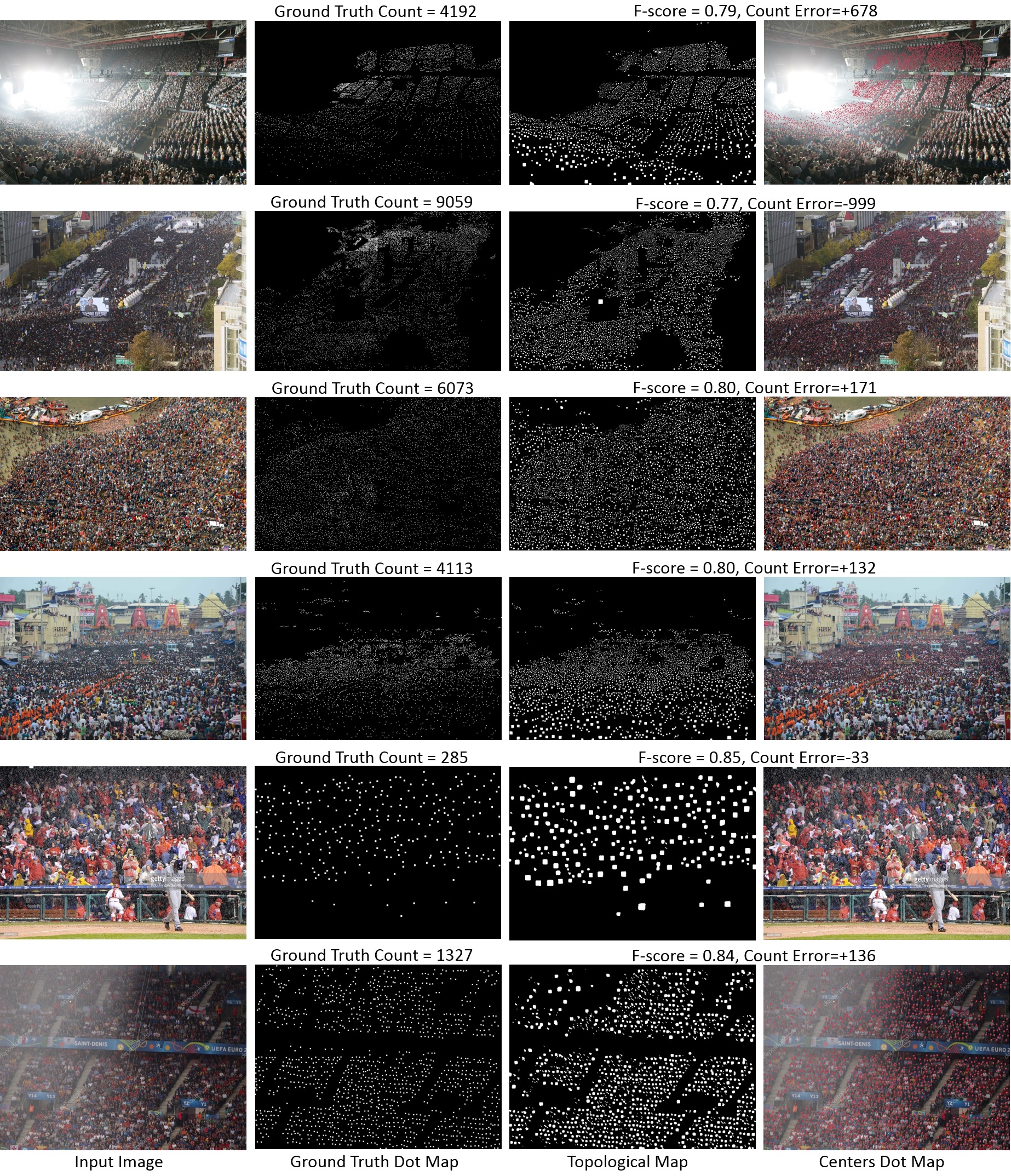}
\caption{Sample results from some difficult cases in the JHU++ crowd counting dataset. The columns from left to right are: the input image, the ground truth dot map, the predicted topological map, and the centers of the components in the topological map overlayed on the input image as red dots.}
\label{fig:jhu_a}
\end{figure*}

\begin{figure*}[htbp]
\centering
\includegraphics[width=1.0\linewidth]{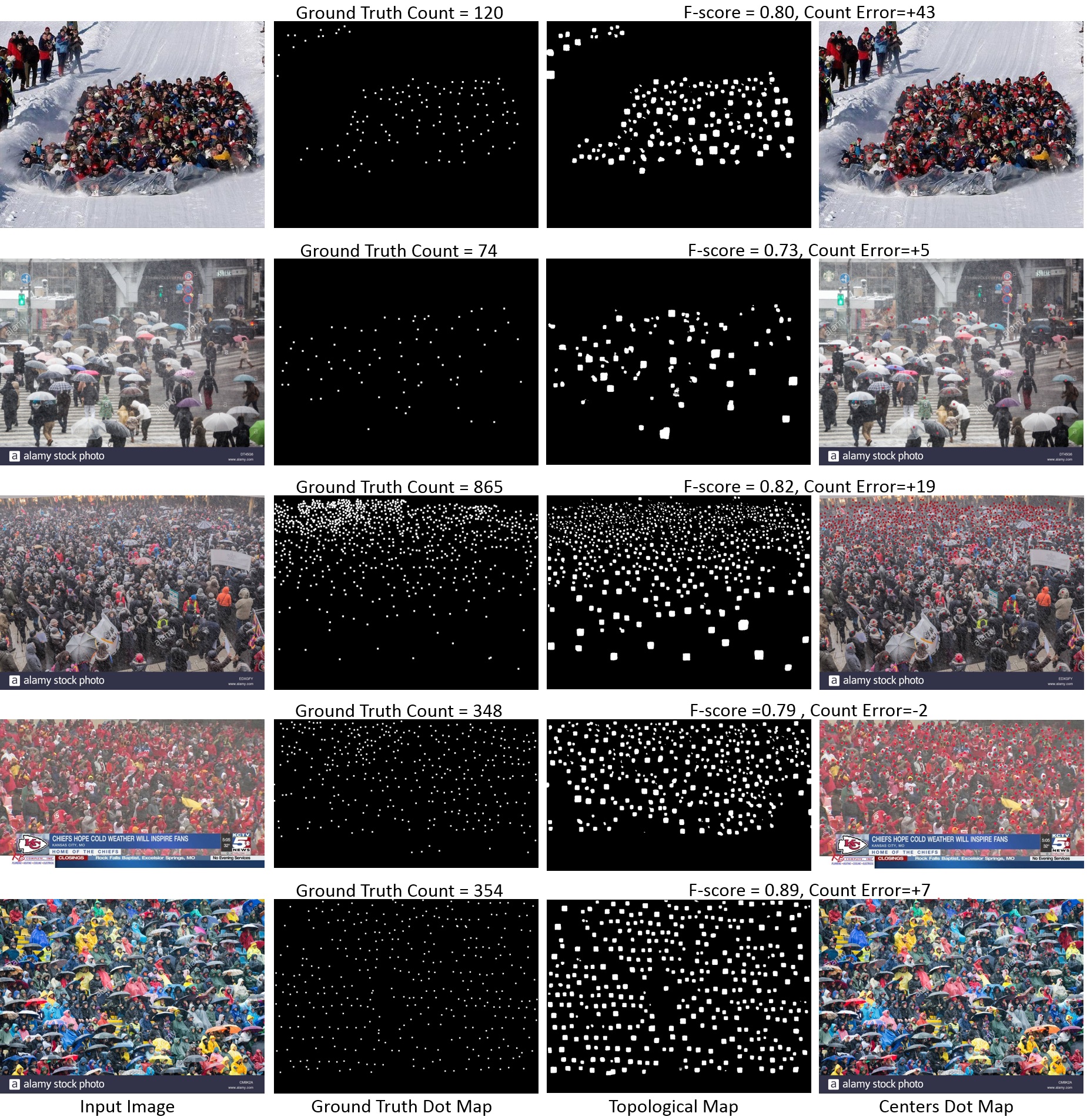}
\caption{Sample results from some difficult cases in the JHU++ crowd counting dataset. The columns from left to right are: the input image, the ground truth dot map, the predicted topological map, and the centers of the components in the topological map overlayed on the input image as red dots. }
\label{fig:jhu_b}
\end{figure*}

The model is trained with crops of 1024x1024 pixels of the original image without any resizing. There are a few images in the training set with no heads at all. We did not use them in training. The dataset contains images with varying difficulties including weather conditions such as rain and fog. We report in Table~\ref{table:jhu-val} and Table~\ref{table:jhu-test} the categorized counting performance on the validation and test sets, respectively. The categories are:
\textit{Low}: images containing count between~0 and~50,
\textit{Medium}: images containing count between~51 and~500,
\textit{High}: images with count more than~500 people,
\textit{Weather}: weather degraded images, and
\textit{Overall}: all the images in the set.
We see that TopoCount and the integration of TopoCount with density-based methods achieve the best performance across most categories. Fig.~\ref{fig:jhu_a} and Fig.~\ref{fig:jhu_b} show sample results of TopoCount on some difficult cases in the JHU++ dataset.

\setlength{\tabcolsep}{4pt}
\begin{table*}[htbp]
\small
\begin{center}
\begin{tabular}{|p{0.08\linewidth}|p{0.09\linewidth}|p{0.09\linewidth}|p{0.075\linewidth}|p{0.09\linewidth}|p{0.09\linewidth}|p{0.075\linewidth}|p{0.09\linewidth}|p{0.09\linewidth}|p{0.075\linewidth}|}

\hline
 & \multicolumn{9}{p{0.81\linewidth}|}{\centering {ShanghaiTech A}}  
 \cr
\cline{2-10}
  & \multicolumn{3}{p{0.255\linewidth}|} {\centering $\sigma=40$}  
  & \multicolumn{3}{p{0.255\linewidth}|} {\centering $\sigma=20$}  
  & \multicolumn{3}{p{0.255\linewidth}|} {\centering $\sigma=5$} 
  \cr
\cline{2-10}
  & \centering {AP.50/AR.50}& \centering {AP.75/AR.75}& \centering {mAP/mAR} 
  & \centering {AP.50/AR.50}& \centering {AP.75/AR.75}& \centering {mAP/mAR} 
  & \centering {AP.50/AR.50}& \centering {AP.75/AR.75}& \centering {mAP/mAR} 
  \cr 
\hline
RAZ\_Loc           
    & \centering $ 74.5/84.7$ & \centering $69.9/82.0$ & \centering $69.1/81.2$ 
    & \centering $ 66.7/79.9$ & \centering $60.1/75.3$ & \centering $58.4/74.1$ 
    & \centering $ 36.0/40.9$ & \centering $20.5/\textbf{57.9}$ & \centering $19.7/42.2$ 
    \cr 
TopoCount 
    & \centering $\textbf{91.0}/\textbf{88.6}$  & \centering $\textbf{89.6}/\textbf{87.2}$ & \centering $\textbf{89.2}/\textbf{86.8}$
    & \centering $\textbf{88.6}/\textbf{86.2}$  & \centering $\textbf{86.1}/\textbf{83.8}$ & \centering $\textbf{85.0}/\textbf{82.8}$
    & \centering $\textbf{72.5}/\textbf{70.9}$  & \centering $\textbf{57.6}/56.4$ & \centering $\textbf{56.0}/\textbf{54.8}$
    \cr 
\hline
 & \multicolumn{9}{p{0.81\linewidth}|}{\centering {UCF QNRF}}  
 \cr
\hline
RAZ\_Loc             
    & \centering $ 57.3/71.9$ & \centering $48.1/65.2$ & \centering $46.2/63.6$ 
    & \centering $ 41.4/60.2$ & \centering $28.7/49.7$ & \centering $28.4/48.3$ 
    & \centering $ 7.9/24.2$ & \centering $3.1/14.3$ & \centering $3.7/14.8$ 
    \cr 
TopoCount 
    & \centering $\textbf{87.5}/\textbf{84.2}$  & \centering $\textbf{83.3}/\textbf{80.2}$  & \centering $\textbf{81.7}/\textbf{78.6}$
    & \centering $\textbf{79.7}/\textbf{76.7}$  & \centering $\textbf{71.1}/\textbf{68.6}$ & \centering $\textbf{69.0}/\textbf{66.5}$
    & \centering $\textbf{41.7}/\textbf{40.3}$  & \centering $\textbf{26.3}/\textbf{25.3}$ & \centering $\textbf{27.1}/\textbf{26.2}$
    \cr 
\hline
\end{tabular}
\caption{Localization accuracy using metric in \cite{RAZ-Net:detect:cvpr:2019}.}
\label{table:raz_metric2}
\end{center}
\vspace{-.15in}
\end{table*}
\setlength{\tabcolsep}{1.4pt}

\setlength{\tabcolsep}{4pt}
\begin{table*}[htbp]
\small
\begin{center}
\begin{tabular}{|p{0.26\linewidth}|p{0.27\linewidth}|p{0.06\linewidth}|p{0.27\linewidth}|p{0.06\linewidth}|}

\hline
& \multicolumn{2}{p{0.33\linewidth}|}{\centering {$\sigma_l$}}  
& \multicolumn{2}{p{0.33\linewidth}|}{\centering {$\sigma_s$}}  
   \\
\cline{2-5}
Method  & 
\centering {A0 / A1 / A2 / A3 / A4 / A5}  
& \centering {Average}  
& \centering {A0 / A1 / A2 / A3 / A4 / A5}  
& \centering {Average}  
\cr
\hline
{Faster RCNN \cite{fasterrcnn:nips:2015}}
    & \centering  0.0 / 00.0 / 00.0 / 07.9 / 37.2 / 63.5
    & \centering 18.2
    & \centering  0.0 / 00.0 / 00.0 / 07.3  / 35.4 / 60.2
    & \centering 17.2
     \cr
\hline
{TinyFaces \cite{tinyfaces:cvpr:2017}} 
    & \centering 4.2 / 22.6 / 59.1 / \textbf{90.0} / \textbf{93.1} / \textbf{89.6}
    & \centering 59.8
    & \centering  3.7 / 19.6 / 54.1 / \textbf{85.8} / \textbf{89.7} / \textbf{84.3}
    & \centering \textbf{56.2}
     \cr
\hline
{VGG+GPR \cite{vgg:grp:gao2019domainadaptive}} 
    & \centering  3.1 / 27.2 / 49.1 / 68.7 / 49.8 / 26.3
    & \centering 37.4
    & \centering  2.7 / 18.6 / 37.8 / 63.0 / 45.7 / 16.1
    & \centering 30.6
     \cr
\hline     

{RAZ\_Loc \cite{RAZ-Net:detect:cvpr:2019} }
    & \centering  5.1 / 28.2 / 52.0 / 79.7 / 64.3 / 25.1
    & \centering 42.4
    & \centering  \textbf{4.6} / 20.7 / 43.3 / 75.2 / 60.2 / 15.9
    & \centering 36.7
    \cr
\hline    
{TopoCount (ours)  }
    & \centering \textbf{5.7} / \textbf{40.6} / \textbf{70.7} / 82.4  / 85.3  / 86.2
    & \centering \textbf{61.8}
    & \centering  \textbf{4.6} / \textbf{28.1} / \textbf{60.3} /  79.0 / 81.2  / 77.5
    & \centering 55.2
    \cr

\hline
\end{tabular}
\caption{NWPU-Crowd Online Localization Challenge. The table reports the recall at different head area ranges and their average. }
\label{table:nwpu2}
\end{center}
\end{table*}
\setlength{\tabcolsep}{1.4pt}

\begin{table*}[htbp]
\begin{center}
\begin{tabular}{|p{0.33\linewidth}|p{0.06\linewidth}|p{0.06\linewidth}|p{0.06\linewidth}|p{0.06\linewidth}|p{0.06\linewidth}|p{0.06\linewidth}|p{0.06\linewidth}|p{0.06\linewidth}|p{0.06\linewidth}|p{0.06\linewidth}|}
\hline
 JHU++ Category & \multicolumn{2}{p{0.12\linewidth}|}{\centering {Low}}  & \multicolumn{2}{p{0.12\linewidth}|}{\centering {Medium}} & \multicolumn{2}{p{0.12\linewidth}|}{\centering {High}} & \multicolumn{2}{p{0.12\linewidth}|}{\centering {Weather}}& \multicolumn{2}{p{0.12\linewidth}|}{\centering {Overall}}\\
\hline
Model & \centering {MAE}  & \centering {RMSE}  & \centering {MAE}  & \centering {RMSE}  & \centering {MAE}  & \centering {RMSE}  & \centering {MAE}  & \centering {RMSE} & \centering {MAE}  & \centering {RMSE}\cr 
\hline 
MCNN \cite{multicolumn:cc:cvpr:2016} & \centering 90.6 & \centering 202.9 & \centering 125.3 & \centering 259.5 & \centering 494.9 & \centering 856.0 & \centering 241.1 & \centering 532.2 & \centering 160.6 & \centering 377.7  \cr 
CSRNet \cite{csr:cvpr:2018} & \centering 22.2 & \centering 40.0 & \centering 49.0 & \centering 99.5 & \centering 302.5 & \centering 669.5 & \centering 83.0 & \centering 168.7 & \centering 72.2 & \centering 249.9 \cr 
SANet \cite{sanet:eccv:2018} & \centering  13.6  & \centering 26.8  & \centering 50.4  & \centering 78.0  & \centering 397.8  & \centering 749.2  & \centering 72.2  & \centering 126.7  & \centering 82.1  & \centering 272.6 \cr 
SFCN \cite{sfcn:cvpr:2019} & \centering  11.8 & \centering 19.8 & \centering 39.3 & \centering 73.4 & \centering 297.3 & \centering 679.4 & \centering \textbf{52.3} & \centering \textbf{93.6} & \centering 62.9 & \centering 247.5 \cr 
LSC-CNN \cite{locate:LSC-CNN:babu:ipmi:2020} & \centering  {6.8} & \centering \textbf{10.1} & \centering 39.2 & \centering 64.1 & \centering 504.7 & \centering 860.0 & \centering 77.6 & \centering 187.2 & \centering 87.3 & \centering 309.0 \cr
JHU++ 
\cite{sindagi2020jhu-crowd++:dataset} & \centering 11.7 & \centering 24.8 & \centering  35.2 & \centering {57.5} & \centering {273.9} & \centering 676.8 & \centering {54.0} & \centering {106.8} & \centering 57.6 & \centering 244.4\cr
CAN \cite{can:cvpr:2019}  & \centering  34.2  & \centering 69.5  & \centering 65.6  & \centering 115.3  & \centering 336.4  & \centering {619.7}  & \centering 101.8  & \centering 179.3  & \centering 89.5  & \centering 239.3 \cr 
Bayesian \cite{bayesian:2019:ICCV}  & \centering 6.9 & \centering {10.3} & \centering 39.7 & \centering 85.2 & \centering 279.8 & \centering 620.4 & \centering 58.9 & \centering 124.7 & \centering 59.3 & \centering 229.2\cr 
CAN+TopoCount (\textbf{ours})& \centering 26.5 & \centering 49.9 & \centering 38.2 & \centering 63.3 & \centering 277.1 & \centering 621.4 & \centering 62.5 & \centering 112.0 & \centering 64.9 & \centering {227.3}\cr 
Bayesian + TopoCount (\textbf{ours})& \centering \textbf{6.3} & \centering 11.0 & \centering {32.5} & \centering 58.8 & \centering \textbf{269.6} & \centering \textbf{602.1} & \centering 62.6 & \centering 123.7 & \centering \textbf{54.1} & \centering \textbf{218.1}\cr 
TopoCount (\textbf{ours}) & \centering 6.9 & \centering 11.1 & \centering \textbf{32.1} & \centering \textbf{51.8} & \centering 275.7 & \centering 633.8 & \centering 57.5 & \centering 119.3 & \centering {54.3} & \centering {228.2}\cr 
\hline
\end{tabular}
\caption{Categorical Counting Results On JHU-CROWD++ Dataset (“\textbf{Val Set}”). The descriptions of the categories are in Appendix~\ref{sec:appendix-exp-jhu}}
\label{table:jhu-val}
\end{center}
\end{table*}
\setlength{\tabcolsep}{1.4pt}

\begin{table*}
\begin{center}
\begin{tabular}{|p{0.33\linewidth}|p{0.06\linewidth}|p{0.06\linewidth}|p{0.06\linewidth}|p{0.06\linewidth}|p{0.06\linewidth}|p{0.06\linewidth}|p{0.06\linewidth}|p{0.06\linewidth}|p{0.06\linewidth}|p{0.06\linewidth}|}
\hline
 JHU++ Category & \multicolumn{2}{p{0.12\linewidth}|}{\centering {Low}}  & \multicolumn{2}{p{0.12\linewidth}|}{\centering {Medium}} & \multicolumn{2}{p{0.12\linewidth}|}{\centering {High}} & \multicolumn{2}{p{0.12\linewidth}|}{\centering {Weather}}& \multicolumn{2}{p{0.12\linewidth}|}{\centering {Overall}}\\
\hline
Model & \centering {MAE}  & \centering {RMSE}  & \centering {MAE}  & \centering {RMSE}  & \centering {MAE}  & \centering {RMSE}  & \centering {MAE}  & \centering {RMSE} & \centering {MAE}  & \centering {RMSE}\cr 
\hline
MCNN \cite{multicolumn:cc:cvpr:2016} & \centering  97.1 & \centering 192.3 & \centering 121.4 & \centering 191.3 & \centering 618.6 & \centering 1,166.7 & \centering 330.6 & \centering 852.1 & \centering 188.9 & \centering 483.4 \cr 
CSRNet \cite{csr:cvpr:2018} & \centering  27.1 & \centering 64.9 & \centering 43.9 & \centering 71.2 & \centering 356.2 & \centering 784.4 & \centering 141.4 & \centering 640.1 & \centering 85.9 & \centering 309.2 \cr 
SANet \cite{sanet:eccv:2018} & \centering  17.3 & \centering 37.9 & \centering 46.8 & \centering 69.1 & \centering 397.9 & \centering 817.7 & \centering 154.2 & \centering 685.7 & \centering 91.1 & \centering 320.4 \cr 
SFCN \cite{sfcn:cvpr:2019} & \centering  16.5 & \centering 55.7 & \centering 38.1 & \centering 59.8 & \centering 341.8 & \centering 758.8 & \centering 122.8 & \centering {606.3} & \centering 77.5 & \centering 297.6 \cr 
LSC-CNN \cite{locate:LSC-CNN:babu:ipmi:2020} & \centering  10.6 & \centering 31.8 & \centering 34.9 & \centering 55.6 & \centering 601.9 & \centering 1,172.2 & \centering 178.0 & \centering 744.3 & \centering 112.7 & \centering 454.4 \cr
JHU++
\cite{sindagi2020jhu-crowd++:dataset} & \centering 14.0 & \centering 42.8 & \centering  35.0 & \centering 53.7 & \centering 314.7 & \centering 712.3 & \centering \textbf{120.0} & \centering \textbf{580.8} & \centering 71.0 & \centering 278.6\cr
CAN \cite{can:cvpr:2019}  & \centering  37.6 & \centering 78.8 & \centering 56.4 & \centering 86.2 & \centering 384.2 & \centering 789.0 & \centering 155.4 & \centering 617.0 & \centering 100.1 & \centering 314.0 \cr 
Bayesian \cite{bayesian:2019:ICCV} & \centering 10.1 & \centering 32.7 & \centering 34.2 & \centering 54.5 & \centering 352.0 & \centering 768.7 & \centering 140.1 & \centering 675.7 & \centering 75.0 & \centering 299.9\cr 
CAN+TopoCount (\textbf{ours})& \centering 30.7 & \centering 60.3 & \centering 38.3 & \centering 63.9 & \centering \textbf{275.0} & \centering \textbf{659.7} & \centering 123.2 & \centering 625.7 & \centering 71.9 & \centering \textbf{260.9}\cr 
Bayesian+TopoCount (\textbf{ours})& \centering \textbf{7.8} & \centering {22.8} & \centering \textbf{28.5} & \centering {52.5} & \centering {286.8} & \centering {670.6} & \centering 122.9 & \centering 639.2 & \centering {61.8} & \centering {262.0}\cr 
TopoCount (\textbf{ours})& \centering {8.2} & \centering \textbf{20.5} & \centering {28.9} & \centering \textbf{50.0} & \centering {282.0} & \centering {685.8} & \centering {120.4} & \centering 635.1 & \centering \textbf{60.9} & \centering {267.4}\cr 
\hline
\end{tabular}
\caption{
Categorical Counting Results On JHU-CROWD++ Dataset (“\textbf{Test Set}”). The descriptions of the categories are in Appendix~\ref{sec:appendix-exp-jhu}}
\label{table:jhu-test}
\end{center}
\end{table*}
\setlength{\tabcolsep}{1.4pt}

